\documentclass[accepted]{uai2023} %

\usepackage{commands}

\title{\titl}

\author[1]{\href{mailto:<sahra.ghalebikesabi@stats.ox.ac.uk>?Subject=Autoregressive Predictive Updates}{Sahra Ghalebikesabi}{}}
\author[2]{Chris Holmes}
\author[2]{Edwin Fong\thanks{equal contribution}}
\author[3]{Brieuc Lehmann$^*$}

\affil[1]{%
    University of Oxford
}
\affil[2]{%
    Novo Nordisk
}
\affil[3]{%
    University College London
  }
  
\begin{document}
\maketitle
\begin{abstract} 
Bayesian methods are a popular choice for statistical inference in small-data regimes due to the regularization effect induced by the prior. %
In the context of density estimation, the standard nonparametric Bayesian approach is to target the posterior predictive of the Dirichlet process mixture model.
In general, direct estimation of the posterior predictive is intractable and so methods typically resort to approximating the posterior distribution as an intermediate step.
The recent development of quasi-Bayesian predictive copula updates, however, has made it possible to perform tractable predictive density estimation without the need for posterior approximation. 
Although these estimators are computationally appealing, they struggle on non-smooth data distributions. This is %
due to the comparatively restrictive form of the likelihood models from which the proposed copula updates were derived. 
To address this shortcoming, we consider a Bayesian nonparametric model with an autoregressive likelihood decomposition and a Gaussian process prior. While the predictive update of such a model is typically intractable, we derive a quasi-Bayesian update that achieves state-of-the-art results in small-data regimes.  
\end{abstract}

\section{Introduction}

Modelling the joint distribution of multivariate random variables with density estimators is a central topic in modern unsupervised machine learning research \citep{durkan2019neural, papamakarios2017masked}.
As well as providing insight into the statistical properties of the data, density estimates are used in a number of downstream applications, including image restoration \citep{zoran2011learning}, density-based clustering \citep{scaldelai2022multiclusterkde}, and simulation-based inference \citep{lueckmann2021benchmarking}. %
In small-data regimes, Bayesian methods are a popular choice for a wide range of machine learning tasks, including density estimation, thanks to their attractive generalization capacities. %
For density estimation, the typical Bayesian approach is to target the \textit{Bayesian predictive density}, 
$
    p_n(x) = \int f(x|\theta) \pi_n(\theta) d\theta, %
$
where $\pi_n$ denotes the posterior density of the model parameters $\theta$ after observing $x_1,\ldots,x_n$, and $f$ denotes the likelihood function. %

De Finetti's representation theorem \citep{de1937prevision, hewitt1955symmetric} states that an exchangeable joint density fully characterises a Bayesian model, which then implies a sequence of predictive densities. Further, \citet{fong2021martingale} recently showed that a sequence of predictive densities can be sufficient for full Bayesian posterior inference. This provides theoretical motivation for an iterative approach to Bayesian predictive density estimation by updating the predictive $p_{i-1}(x)$ to $p_{i}(x)$ given observation $x_i$ for $i=1,\ldots, n$. The idea of recursive Bayesian updates goes back to at least \citet{hill1968posterior}, but {was only recently made more widely applicable} through the relaxation of the assumption of exchangeability in favour of conditionally identically distributed \citep{berti2004limit} sequences.  %
\begin{figure*}
\centering
\subfloat{\includegraphics[width = \textwidth/8]{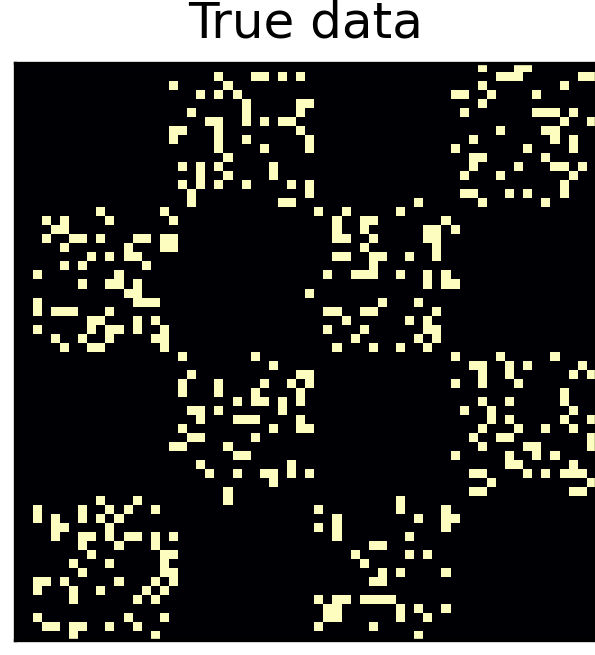}} 
\subfloat{\includegraphics[width = \textwidth/8]{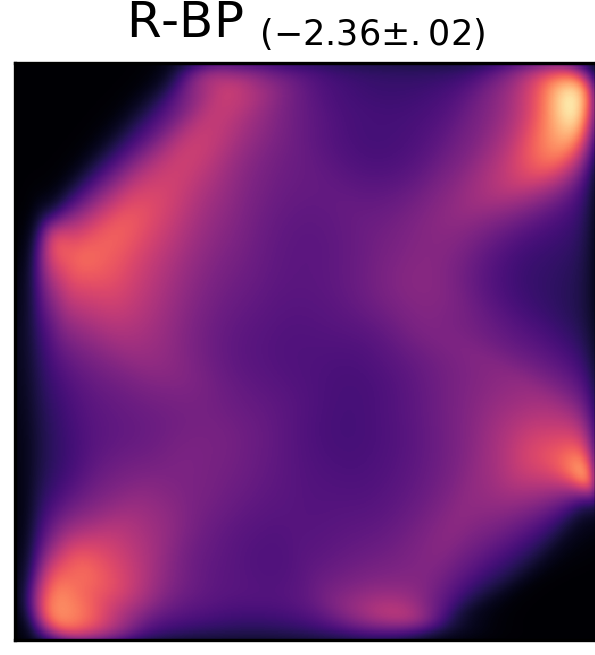}}
\subfloat{\includegraphics[width = \textwidth/8]{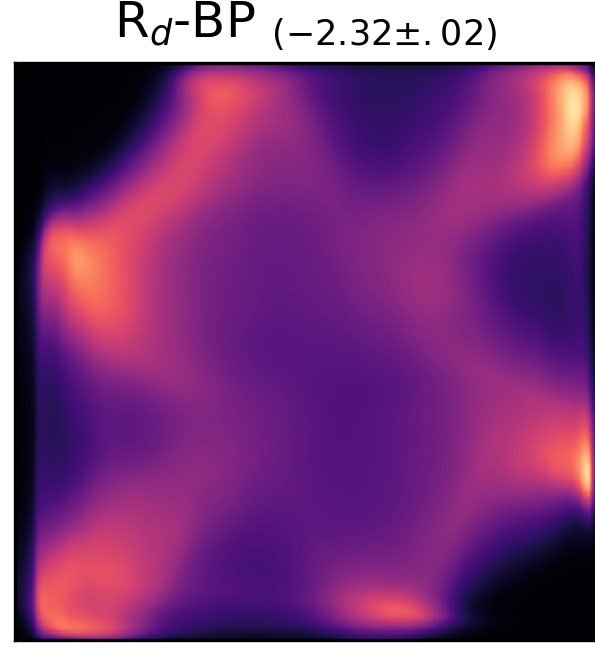}}
\subfloat{\includegraphics[width = \textwidth/8]{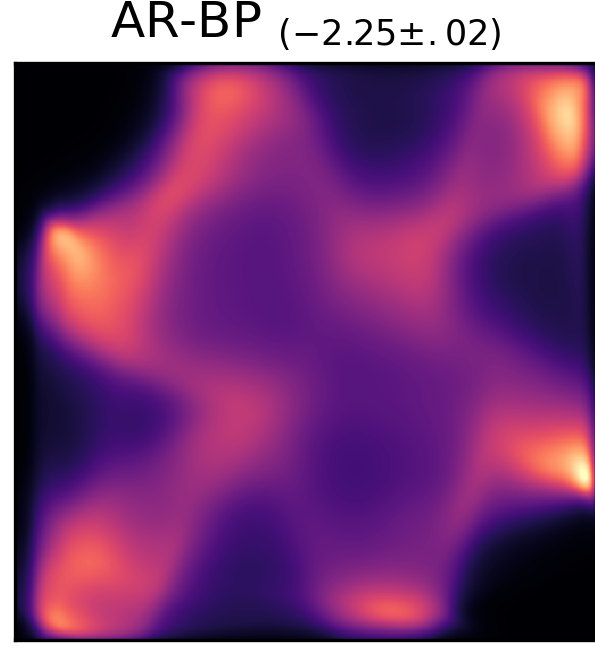}}
\subfloat{\includegraphics[width = \textwidth/8]{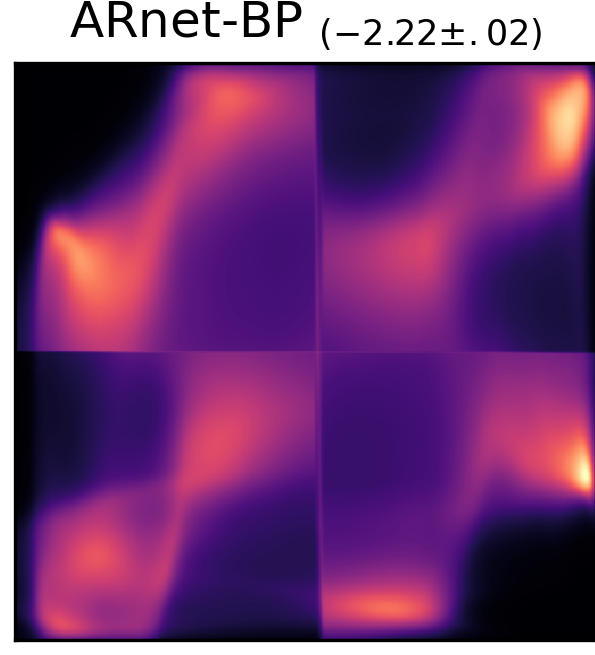}} 
\subfloat{\includegraphics[width = \textwidth/8]{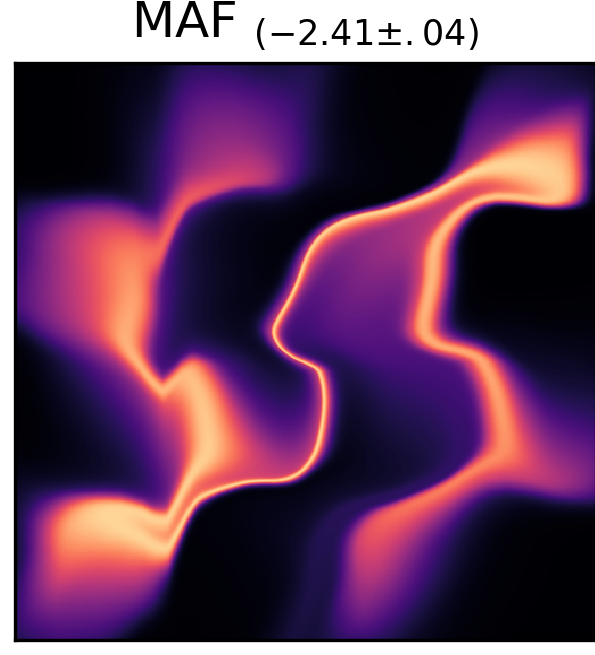}} 
\subfloat{\includegraphics[width = \textwidth/8]{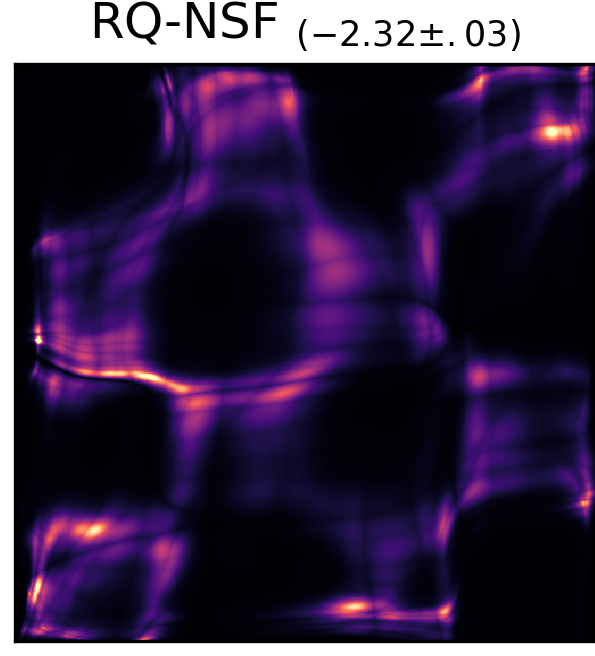}} 
\caption{Density estimates of 600 observations from a chessboard distribution, reported with mean and standard deviation of test log likelihoods. For larger training sizes, see Supplement \ref{app:exp-toy}. Our methods, AR-BP and ARnet-BP, outperform R-BP and AR neural networks.}
\label{fig:chess}
\end{figure*}

Here, we focus on a particular class of one-step-ahead predictive updates $p_{i-1}(x)\rightarrow p_{i}(x)$ based on bivariate copulas, which were first introduced by \citet{hahn2018recursive} for univariate data, and extended by \citet{fong2021martingale} to the multivariate setting and to regression analyses. This class of updates is {inspired} by Bayesian models and thus retains many desirable Bayesian properties, such as coherence and regularization. However, we emphasize that the copula updates do not correspond exactly, nor approximately, to a traditional Bayesian likelihood-prior model, and we thus refer to them as \textit{quasi-Bayesian} \citep{fortini2020quasi}.  %
The most related Bayesian density estimator proposed to date, henceforth referred to as the \textbf{R}ecursive \textbf{B}ayesian \textbf{P}redictive (R$_d$-BP), lacks flexibility to model highly complex data distributions (see Figure \ref{fig:chess}).
This is because the existing copula updates rely on a Gaussian copula with a single scalar bandwidth parameter, corresponding to a Bayesian model with a likelihood that factorizes over dimensions. In contrast, popular neural network based approaches, such as \glspl{maf} \citep{papamakarios2017masked}, and \glspl{rqnsf} \citep{durkan2019neural} can struggle in small-data regimes (see Figure \ref{fig:chess}).

\paragraph{Contributions} This motivates our main contribution, namely the formulation of a more flexible \gls{ar} copula update based on which we propose a new \gls{dpmm} inspired density estimator. In particular:
\begin{itemize}
    \item By considering a \gls{dpmm} with an \gls{ar} likelihood and a \gls{gp} prior, we formulate a tractable copula update with a novel \textit{data-dependent bandwidth} based on the Euclidean metric in data space.
    Our method, \textbf{A}utoregressive \textbf{R}ecursive \textbf{B}ayesian \textbf{P}redictives (AR-BP), outperforms traditional density estimators on tabular data with up to 63 features, and 10,000 samples. %
    \item We observe in practice that the Euclidean metric used in AR-BP can be inadequate for highly non-smooth data distributions. For such cases, we propose using an \gls{ar} neural network \citep{bengio1999modeling, frey1998graphical, germain2015made, larochelle2011neural} that maps the observations into a latent space before bandwidth estimation. This introduces additional non-linearity through the dependence of the bandwidth on the data, leading to a density estimator, ARnet-BP, that is more accurate on non-smooth densities.
\end{itemize}
\section{Background} 

We briefly recap predictive density estimation via bivariate copula updates%
, before describing a particular such update inspired by \glspl{dpmm}.

\subsection{Univariate Predictive Density Updates}

To compute predictive densities quickly, \citet{hahn2018recursive} propose an iterative approach.
For $x \in \mathbb{R}$, any sequence of Bayesian posterior predictive densities $p_i(x)$ with likelihood $f$ and posterior $\pi_i$, conditional on  $x_{1:i}$, can be expressed as 
\begin{align}
    {p_i(x)} =\int f(x|\theta) \pi_i(\theta) d\theta = {p_{i-1}(x)}h_i(x, x_i), \label{eq:update}
\end{align}
for some bivariate function $h_i(x, x_i)$ \citep{hahn2018recursive}. Rearranging for $h_i$, we have
\begin{align}
    h_i(x, x_i){=}\!\frac{p_{i}(x)}{p_{i-1}(x)}  \!\!\overset{(a)}{=} \!\! \frac{p_{i-1}(x|x_i)}{p_{i-1}(x)} 
     \!\!\overset{(b)}{=}\!\!
    \frac{p_{i-1}(x, x_i)}{p_{i-1}(x)p_{i-1}(x_i)}
    \label{eq:h}
\end{align}
where (a) holds by definition, and (b) $p_{i-1}(x,x_i) = p_{i-1}(x|x_i)p_{i-1}(x_i) = p_i(x)p_{i-1}(x_i)$ holds by Bayes' law. \citet{hahn2018recursive} show that $h_i(x, x_i)$ is the transformation of a bivariate copula density.
A \textit{bivariate copula} is a bivariate \gls{cdf} $C:[0, 1]^2\rightarrow [0, 1]$ with uniform marginal distributions that is used to characterise the dependence between two random variables independent of their marginals:

\begin{thm}[Sklar's theorem \citep{sklar1959fonctions}]
For any bivariate density $f(y_1, y_2)$ with continuous marginal \glspl{cdf}, $F_1(y_1)$ and $F_2(y_2)$, and marginal densities $f_1(y_1)$ and $f_2(y_2)$, there exists a unique bivariate copula $C$ with density $c$ such that
\begin{align*}
    f(y_1, y_2) = c\left\{F_1(y_1),F_2(y_2)\right\}f_1(y_1)f_2(y_2).
\end{align*}
\end{thm}
Applying the copula factorization from Sklar's theorem to \eqref{eq:h} yields that there exists some bivariate copula density $c_i$ such that
$p_{i-1}(x, x_i)=c_i\{P_{i-1}(x), P_{i-1}(x_i)\}p_{i-1}(x)p_{i-1}(x_i),$
and thus
$
    h_i(x,x_i) = c_i\{P_{i-1}(x), P_{i-1}(x_i)\},
$
where $P_{i-1}$ is the \gls{cdf} corresponding to the predictive density $p_{i-1}$. Given prior $\pi$ and likelihood $f$, Equation \ref{eq:h} suggests that the update function can be written as
\begin{align*}
    h_i(x, x_i) = \frac{\int f(x|\theta)f(x_i|\theta)\pi_{i-1}(\theta) d\theta}{\int f(x|\theta)\pi_{i-1}(\theta)d\theta\int f(x_i|\theta)\pi_{i-1}(\theta)d\theta}\cdot %
\end{align*}
For each Bayesian model, there is thus a unique sequence of symmetric copula densities $c_i(u, v) = c_i(v, u)$.
This sequence has the property that $c_n(\cdot, \cdot)\rightarrow 1$ converges to a constant function as $n\rightarrow \infty$, ensuring that the predictive density converges asymptotically with sample size $n$. 

In general, the above equation is intractable due to the posterior so it is not possible to compute the iterative update in \eqref{eq:update} for fully Bayesian models. %
Alternatively, we will consider sequences of $h_i$ that match the Bayesian model for $i = 1$, but not for $i>1$.
As mentioned above, this copula update no longer corresponds to a Bayesian model, nor are the resulting predictive density estimates approximations to a Bayesian model. Nevertheless, if the copula updates are \textit{conditionally identically distributed}, they still exhibit desirable Bayesian characteristics such as coherence and regularization, and are hence referred to as \textit{quasi-Bayes}.
Please refer to \cite{berti2004limit%
} for details.

\subsection{Multivariate Predictive Density Updates}  
The above arguments cannot directly be extended to multivariate $x \in \mathbb{R}^d$ since $h_i$ cannot necessarily be written as $c_i\{P_{i-1}(x), P_{i-1}(x_i)\}$ for $d > 1$. However, \eqref{eq:h} still holds, and recursive predictive updates with bivariate copulas as building blocks can be derived explicitly given a pre-defined likelihood model and a prior, which we now exhibit.

\citet{hahn2018recursive} and \citet{fong2021martingale} propose to use \glspl{dpmm} as a general-use nonparametric model. The \gls{dpmm} \citep{escobar1988estimating, escobar1995bayesian}
can be written as
\begin{align}
    f(x| G) = \int_{\Theta} K(x|\theta) \, dG(\theta), \text{ with }G\sim \text{DP}(c, G_0) \label{eq:DPlh}
\end{align}
where $\theta \in \Theta=\mathbb{R}^d$ are parameter vectors, the prior assigned to $G$ is a \gls{dp} prior with base measure $G_0$ and concentration parameter $c>0$ \citep{ferguson1973bayesian}, and $K(x|\theta)$ is a user-specified kernel (not to be confused with the covariance function of a \gls{gp}).
In particular, \citet{fong2021martingale} consider the base measure $G_0=\Normal(0, \tau^{-1}I_d)$ for some precision parameter $\tau\in\mathbb{R}_{>0}$, and the factorized kernel $K(x|\theta)=\Normal(x|\theta, I_d)$ where $I_d$ is the $d$-dimensional identity matrix. The likelihood is then %
\begin{align}
f(x| G) = \int \prod_{j=1}^d\Normal\left(x^j \mid \theta^j,1\right) dG({\theta}),\label{eq:mv_fact_DP_mixture}
\end{align}
where the dimensions of $x$ are conditionally independent given $\theta$. Following \cite{hahn2018recursive}, we denote the dimension $j$ of a vector $y$ with $y^{j}$. We note that the strong assumption of a factorised kernel form drastically impacts the performance of the regular \gls{dpmm} and also influences the form and modelling capacity of the corresponding copula update.

This model inspires the following recursive predictive density update $p_{i}(x) = h_i(x, x_i)p_{i-1}(x)$ for which the first $d'\in\{1,\ldots,d\}$ marginals take on the form
\begin{align}\label{eq:mv_DP_marginal} %
\frac{p_{i}({x}^{1:d'})}{p_{i-1}\left(x^{1:d'}\right)}\!=& %
1\!-\!\alpha_{i}\!
+\!\alpha_{i}\! \prod_{j=1}^{d'} \!c\left(u_{i-1}^{j}(x^j),v_{i-1}^{j}; \rho_0 \right)
, \\
u_{i-1}^{j}(x^j) :=& P_{i-1}\left(x^j \mid  x^{1:j-1}\right),\nonumber \\ 
v_{i-1}^{j} :=& P_{i-1}\left(x_{i}^j \mid x_{i}^{1:j-1}\right), \nonumber 
\end{align}
 where $c(u,v; \rho_0)$ is the bivariate Gaussian copula density with correlation $\rho_0 = 1/(1+\tau)$, { $p_0$ can be any chosen prior density},  and $\alpha_i=\left(2-\frac{1}{i}\right)\frac{1}{i+1}$ (see Supplement A and \cite{fong2021martingale}). {Note that the above update requires a specific ordering of the feature dimensions, and the Gaussian copula follows from the Gaussian distribution in the kernel and $G_0$ for the \gls{dpmm}}. %
Unlike the DPMM, there are now no underlying parameters (beyond $\rho_0$) in the copula update as we have integrated out $\theta$, so we do not carry out clustering directly.
While $\rho_0$ is a scalar here, \citet{fong2021martingale} also consider the setting with a distinct bandwidth parameter for each dimension. We refer to these recursive Bayesian predictives as R$_d$-BP, or simply R-BP if the dimensions share a single bandwidth.

\section{AR-BP: Autoregressive Bayesian Predictives}
\label{sec:meth}

For smooth data distributions, the recursive update defined in \eqref{eq:mv_DP_marginal} generates density estimates that are highly competitive against other popular density estimation procedures such as \gls{kde} and \gls{dpmm} \citep{fong2021martingale}. Moreover, the iterative updates provide a fast estimation alternative to fitting the full \gls{dpmm} through \gls{mcmc}. When considering more structured data, however, performance suffers due to the choices of the factorized kernel $K(\cdot|\theta)=\Normal(\cdot|\theta, I_d)$ and simple base measure $G_0=\Normal(0, \tau^{-1}I_d)$ in the \gls{dpmm}. These choices induce a priori independence between the data dimensions, and are thus insufficiently flexible to capture more complex dependencies. 

\subsection{Bayesian Model Formulation}

We therefore propose employing more general kernels and base measures in the \gls{dpmm} and show that these inspire a more general tractable recursive predictive update. In particular, we allow the kernel to take on an autoregressive structure
\begin{align} \label{eq:mv_AR_DP_kernel}
K(x|\theta) = \prod_{j=1}^d\Normal\left(x^j \mid \theta^j\left(x^{1:j-1}\right),1\right),
\end{align}
where $\theta^j:\mathbb{R}^{j-1}\rightarrow \mathbb{R}$ is now an unknown mean \textit{function}, and not scalar, for dimension $x^j$, which we allow to depend on the previous $j-1$ dimensions of $x$. Thus, specifying our \gls{dpmm} requires a base measure supported on the function space in which $(\theta^1,\dots,\theta^d)$ is valued. We specify this base measure as a product of independent \gls{gp} priors on the functional parameters %
\begin{align} \label{eq:mv_AR_DP_base}
    \theta^j\sim \text{GP}(0, \tau^{-1} k^j) \text{ for } j=1,...,d
\end{align}
where $k^j : \mathbb{R}^{j-1} \times \mathbb{R}^{j-1}\rightarrow \mathbb{R}$ 
and $k^j$ can be any given covariance function that takes as input a pair of ${x}^{1:j-1}$ values.  In practice, we use the same functional form of $k$ for each $j$, so we will drop the superscript $j$. For later convenience, we have also written the scaling term $\tau^{-1}$ explicitly. We highlight that for $j=1$, $\theta^1 \sim \Normal(0, \tau^{-1})$.
Under this choice, the mean of the normal kernels in the \gls{dpmm} for each dimension $j$ is thus a flexible function of the first $j-1$ dimensions $x^{1:j-1}$, on which we elicit independent \gls{gp} priors. The conjugacy of the \gls{gp} with the Gaussian \gls{dpmm} kernel in \eqref{eq:mv_AR_DP_kernel} is crucial for deriving a tractable density update.

\begin{rem}
The proposed \gls{dpmm} kernel in \eqref{eq:mv_AR_DP_kernel} is in fact more flexible than a general multivariate kernel, $K(x \mid \theta) = \Normal(x \mid \theta, \Sigma)$. This is because the multivariate kernel also implies an \gls{ar} form like \eqref{eq:mv_AR_DP_kernel} but where the parameters $\theta^j$ are restricted to be linear in $x^{1:j-1}$; see \cite{wade2014predictive} for details.
\end{rem}

\subsection{Iterative Predictive Density Updates}
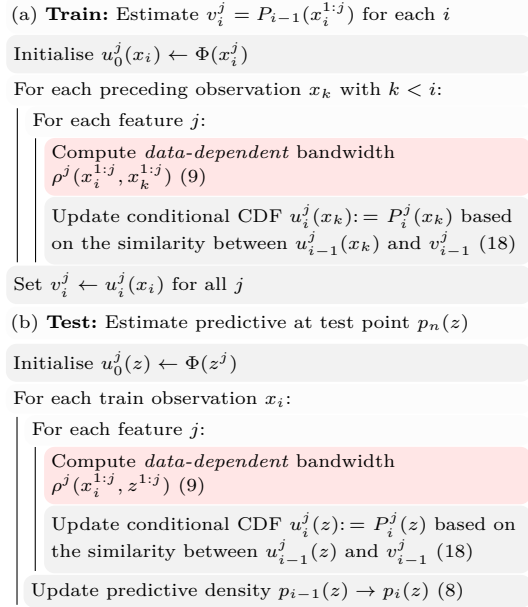
\begin{figure}
\scriptsize
\begin{subfigure}[t]{0.5\textwidth}
\centering
\begin{tikzpicture}%
\node (goal) [fornode0] at (-1,1.95) {(a) \textbf{Train:} Estimate $v_i^j
=P_{i-1}(x^{1\colon j}_i)
$ 
for each $i$
};
\node (init) [wiiidenode] at (-1,1.42) {Initialise ${u_{0}^j(x_{i})} \gets \Phi(x_{i}^j)$};
\node (forobs) [fornode0] at (-1,0.94) {For each preceding observation $x_k$  with $k<i$:};
\node (forfeat) [fornode1] at (-0.875,0.52) {For each feature $j$:};
\node (bandwidth) [newnode] at (-0.75, -0.08) {Compute \textit{data-dependent} bandwidth\\ $\rho^j(x^{1\colon j}_i, x^{1\colon j}_k)$ 
\eqref{eq:rho}};
\node (cdfupdate) [oldnode] at (-0.75,-0.93) {Update conditional CDF $u_i^j(x_k)\colon=P_i^j(x_k)$ based on the similarity between $u_{i-1}^j(x_k)$ and $v_{i-1}^j$ \eqref{eq:mv_autoreg_DP_copdistr}};
\node (prequpdate) [wiiidenode] at (-1,-1.65) {Set $v_{i}^j\leftarrow u_{i}^j(x_i)$ for all $j$};
\draw (-4.31, 0.7) -- (-4.31, -1.35); 
\draw (-4.07, 0.3) -- (-4.07, -1.35); 
\end{tikzpicture}
\end{subfigure}%
\hfill
\begin{subfigure}[t]{0.5\textwidth}
\centering
\begin{tikzpicture}%
\node (goal) [fornode0] at (-1,1.95) {(b) \textbf{Test:} Estimate predictive at test point $p_{\nr}(z)$};
\node (init) [wiiidenode] at (-1,1.42) {Initialise ${u_{0}^j(z)} \gets \Phi(z^j)$};
\node (forobs) [fornode0] at (-1,0.94) {For each train observation $x_i$:};
\node (forfeat) [fornode1] at (-0.88,0.52) {For each feature $j$:};
\node (bandwidth) [newnode] at (-0.75,-0.08) {Compute \textit{data-dependent} bandwidth\\$\rho^j(x^{1\colon j}_i, z^{1\colon j})$ \eqref{eq:rho}};
\node (bandwidth) [oldnode] at (-0.75,-0.93) {Update conditional CDF $u_i^j(z)\colon=P_i^j(z)$ based on the similarity between $u_{i-1}^j(z)$ and $v_{i-1}^j$ \eqref{eq:mv_autoreg_DP_copdistr}};
\node (bandwidth) [widenode] at (-0.88,-1.62) {Update predictive density $p_{i-1}(z)\rightarrow p_i(z)$ \eqref{eq:mv_DP_AR_marginal}};
\draw (-4.31, 0.7) -- (-4.31, -1.8); 
\draw (-4.07, 0.3) -- (-4.07, -1.35); 
\end{tikzpicture}
\end{subfigure}%

\caption{Simplified summary of AR-BP. We repeat the training update for each train datum $x_i$ to estimate $v_i^j=P_{i-1}(x^{1:j}_i)$. These are needed at test time to update from $p_{i-1}(z)\rightarrow p_i(z)$. All steps are averaged over different feature and sample permutations. The main step that induces autoregression in the observations is highlighted \textcolor{pink}{pink}. Please see Supplement \ref{app:imp} for detailed algorithms.}
    \label{fig:flow}
\end{figure}

Computing the Bayesian posterior predictive density induced by the \gls{dpmm} with kernel given by \eqref{eq:mv_AR_DP_kernel} and base measure given by \eqref{eq:mv_AR_DP_base} through posterior estimation is \textit{intractable} and requires \gls{mcmc}. However, as before, we can utilize the model to derive tractable iterative copula updates.
In Supplement \ref{app:ar-deriv}, we derive the corresponding recursive predictive density update $p_{i}(x) = h_i(x, x_i)p_{i-1}(x)$ for the first $d'$ marginals and show that it takes on the form
\begin{align}\label{eq:mv_DP_AR_marginal} %
\frac{p_{i}({x}^{1:d'})}{p_{i-1}(x^{1:d'})} =& 1-\alpha_{i} +\alpha_{i}\cdot\\ &\prod_{j=1}^{d'} c\left(u_{i-1}^{j}(x^{j}),v_{i-1}^{j}; \rho^j(x^{1:j-1}, x_i^{1:j-1})\right), \nonumber
\end{align}
with $u_{i-1}^j(x^j), v_{i-1}^j$ defined as in \eqref{eq:mv_DP_marginal}, 
 $\alpha_i=\left(2-\frac{1}{i}\right)\frac{1}{i+1}$, and the bandwidth given by
\begin{equation}\label{eq:rho}
\rho^j(x^{1:j-1}, x_i^{1:j-1}) = \rho_0 k\left(x^{1:j-1}, x_{i}^{1:j-1}\right),
\end{equation}
for $\rho_0 = 1/(1+\tau)$, and $\rho_{i}^1=\rho_0$. Where appropriate, we henceforth drop the argument $x$ for brevity. 
The conditional \gls{cdf}s $u_{i-1}^j$ can also be computed through an iterative closed form expression similarly to \eqref{eq:mv_DP_AR_marginal} (Supplement \ref{app:imp}). Please see Figure \ref{fig:flow} for a simplified overview of the density estimation pipeline.

{Note that the estimation is identical to the update given in \eqref{eq:mv_DP_marginal} induced by the factorized  \gls{dpmm} kernel, except for the main difference that the bandwidth $\rho$ is \textit{no longer a constant}, but is now \textit{data-dependent}.} More precisely, the bandwidth for dimension $j$ is a transformation of the \gls{gp} covariance function $k$ on the first $j - 1$ dimensions. 
The additional flexibility afforded by the inclusion of $k$ enables us to capture more complex dependency structures, as we do not enforce a-priori independence between the dimensions of the parameter $\theta$. Similarly to the extension of R-BP to 
R$_d$-BP, we can also 
define AR$_d$-BP by introducing dimension dependence in $\rho_0$. Finally, we highlight that extending R-BP to mixed data is possible as given in Appendix E.1.3 of \citet{fong2021martingale}, which also extends naturally to AR-BP.
\begin{rem}
The data-dependent bandwidth also appears when starting from other Bayesian nonparametric models, such as dependent \gls{dp}s and \glspl{gp} (see Supplement \ref{app:gp} for the derivation).
\end{rem}

\begin{figure*}
    \centering
    \subfloat{\includegraphics[height=1.65in]{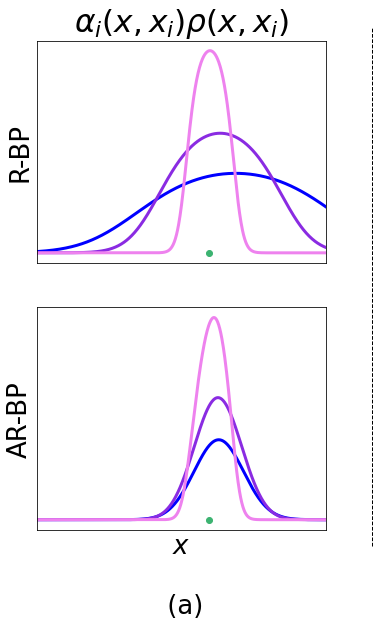}} 
    \hspace{1.15mm}
    \subfloat{\includegraphics[height=1.65in]{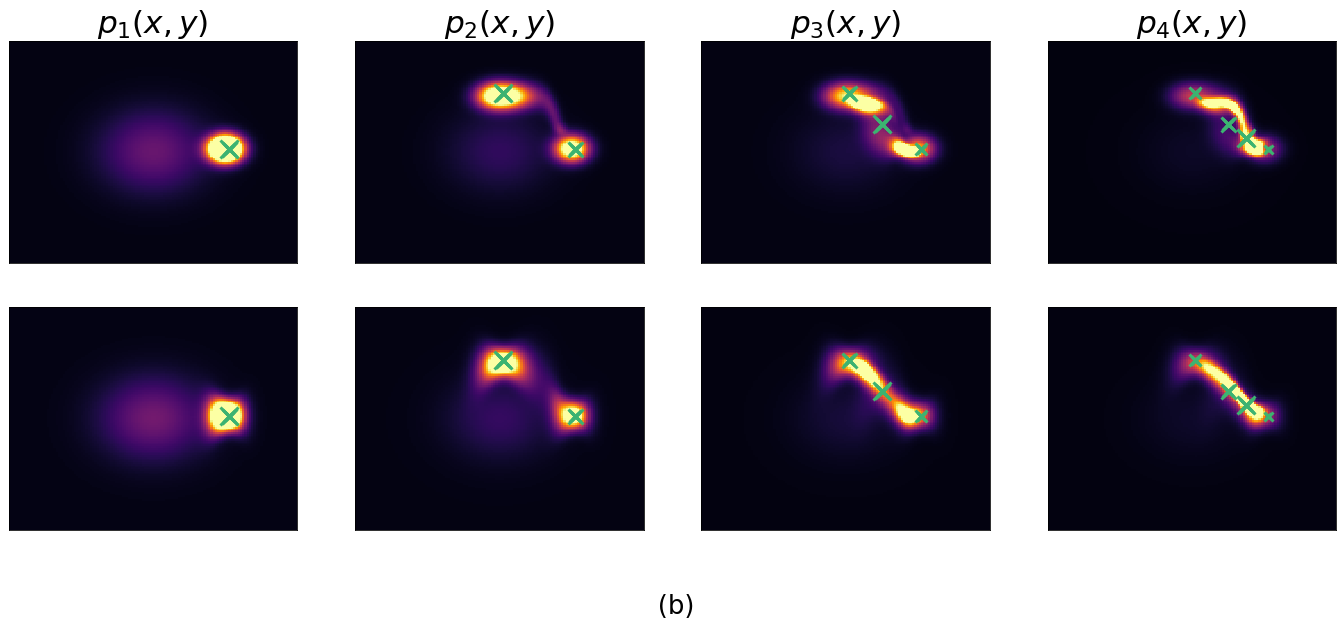}} 
    \caption{(a) Plots of $\alpha_i(x,x_i) \rho(x, x_i)$ for R-BP and AR-BP for $\rho_0\in\{0.5,0.7,0.95\}$ (\full,\fullmid,\fulllow) with new observation $x_i$ (\dotmid). Note that $\rho(x, x_i) = \rho_0$ for R-BP, and $\ell = 1$ for AR-BP.  (b) Density plots for R-BP and AR-BP trained on 4 sequential data points (\crossmid). Both figures show that the update of R-BP, unlike AR-BP, is not centred around the new datum.
    }
    \label{fig:itupd}
\end{figure*}

Our approach can be viewed as a Bayesian version of an online \gls{kde} procedure. To see this, note that a \gls{kde} trained on $i-1$ observations -- yielding the density estimate $q_{i-1}(x)$ -- can be updated after observing the $i^{th}$ observation $x_i$ via $q_{i}(x) = (1-\alpha_i)q_{i-1}(x) + \alpha_i d\left(x, x_{i}\right)$, where $\alpha_i=1/i$ and $d(\cdot,\cdot)$ denotes the kernel of the \gls{kde}. Rather than adding a weighted kernel term directly, AR-BP instead adds an adaptive kernel that depends on a notion of distance between $x$ and $x_i$ based on the  predictive \gls{cdf}s conditional on $x_{1:i-1}$. 

To better understand the importance of the data-dependent bandwidth, we compare the conditional predictive mean of R-BP and AR-BP in the bivariate setting $X \times Y$. Under the simplifying assumption of Gaussian predictive densities, we show in Supplement \ref{app:cond_mean}  that the conditional mean of $Y \mid X$ is given by
\begin{align*}
    \mu_i(x) &= \mu_{i-1}(x) +  \alpha_i(x,x_i) \rho(x, x_i) (y_i - \mu_{i-1}(x_i)),\\
    \alpha_i(x,x_i) &= \frac{\alpha_i c(P_{i-1}(x), P_{i-1}(x_i); \rho)}{1-\alpha_i + \alpha_i c(P_{i-1}(x), P_{i-1}(x_i); \rho)}.
\end{align*}
Note that $\rho(x, x_i)=\rho_0$ for R-BP. Intuitively, the updated mean is the previous mean plus a residual term at $y_i$ scaled by some notion of distance between $x$ and $x_i$. For R-BP, this distance between $x$ and $x_i$ depends only on their predictive \gls{cdf} values through $\alpha_i(x,x_i)$. This can result in undesirable behaviour {as shown in the upper plot in Figure \ref{fig:itupd}(a), where the peak of $\alpha_i(x,x_i)$, as a function of $x$, is not centred at $x_i$.} Counterintuitively, there is thus an $x > x_i$ where $\mu_i(x)$ is updated more than at the actual observed $x = x_i$. This follows from the lack of focus on \textit{conditional} density estimates for R-BP, which is alleviated by AR-BP.
In the \gls{ar} case, $\rho(x, x_i)$ takes into account the Euclidean distance between $x$ and $x_i$ in the data space. We see in the lower plot in Figure \ref{fig:itupd}(a) that the peak is closer to $x_i$. Figure \ref{fig:itupd}(b) further demonstrates this difference on another toy example - we see that R-BP struggles to fit a linear conditional mean function for $n = 4$, focussing density in data sparse regions, while AR-BP succeeds to assign significant density only to points on the data manifold. 

\paragraph{Training the update parameters}
In order to compute the predictive density $p_{n}(x^*)$, we require the vector of conditional \gls{cdf}s $[v_1^{j}, \ldots, v_{n-1}^{j}]$ where $v_i^{j} = P_i(x_{i+1}^j \mid x_{i+1}^{1:j-1})$. Given a bandwidth parameterization, obtaining this vector thus amounts to model-fitting, and each $v_i^j$ requires $i-1$ iterations (Supplement \ref{app:imp}), for $i \in \{1,\ldots,n\}$.
We note that the order of samples and dimensions influences the prediction performance in \gls{ar} density estimators \citep{vinyals2015order}. In practice, averaging over different permutations of these improves performance (Supplement \ref{app:ordering}). Full implementation details can be found in Supplement \ref{app:algs}.

\paragraph{Computational complexity} 
The above procedure results in a computational complexity of $\mathcal{O}(Md\nr^2)$ at the training stage where $M$ is the number of permutations. 
At test time, we have already obtained the necessary conditional prequential CDFs $v_n^{j}$ in computing the prequential log-likelihood above.
As a result, we have a computational complexity $\mathcal{O}(Md\nr)$ for each test observation. Note that the introduction of a data-dependent bandwidth does not increase the computational complexity at train or test time relative to R-BP and only adds a negligible factor to the computational time for the calculation of the bandwidth.

\subsection{Bandwidth Parameterisation} 
The choice of covariance function in \eqref{eq:mv_AR_DP_base} provides substantial modelling flexibility in our AR-BP framework. 
Moreover, the additional parameters associated with the covariance function allow us to tune the implied covariance structure according to the observed data.
This formulation enables us to draw upon the rich literature on the choice of covariance functions for Gaussian processes \citep{williams2006gaussian}.
For simplicity we only consider the most popular such choice here, but study the more flexible rational-quadratic covariance in Supplement \ref{app:exp-add}. The \gls{rbf} covariance function is defined as
$
k_\ell(x^{1:j-1},x^{'1:j-1})  = \exp[-\sum_{\kappa=1}^{j-1}\{({x^\kappa - x^{'\kappa}})/{\ell^\kappa}\}^2 ],
$
where $\ell\in\mathbb{R}^{d-1}_{>0}$ is the length scale.

\paragraph{Neural parameterisation} %
As we saw in the motivating example of the density estimation of a chessboard distribution in Figure \ref{fig:chess}, the \gls{rbf} kernel can restrict the capacity of the predictive density update to capture intricate nonlinearities if the training data size is not sufficient. 
While the parameterization of the bandwidth in \eqref{eq:rho} was initially derived via the first predictive update for a \gls{dpmm}, all we require is that the bandwidth function $\rho^j: \mathbb{R}^{j-1} \times \mathbb{R}^{j-1} \to \mathbb{R}$ lies in (0,1).
We would also like $\rho^j(x^{1:j-1}, x^{'1:j-1})$ to take larger values when $x^{1:j-1}$ and $x^{'1:j-1}$ are `close' in some sense.
Motivated by this observation, we now consider more expressive bandwidth functions that can lead to increased predictive performance. 
In particular, we formulate an \gls{ar} neural network ${f}_{w}: \mathbb{R}^{d} \rightarrow \mathbb{R}^{d\times d'}$ for $d'\in\mathbb{N}$ with the property that the $j^{th}$ row of the output depends only on the first $j-1$ dimensions of the input. Let $Z = {f}_{w}\left(x\right)$ and denoting ${z}^{j}$ to be the $j^{th}$ row of the matrix $Z$, the covariance function is then computed as 
$ %
\rho^{j}(x^{1:j-1}, x^{'1:j-1}) = \rho_0 \exp( -\sum_{\kappa=1}^{j-1} \lvert\lvert {{z}}^{\kappa} -  {z}^{'\kappa}\rvert\rvert^2_2 ). 
$ %

Numerous \gls{ar} neural network models have been extensively used for density estimation \citep{dinh2014nice,huang2018neural,kingma2016improved}.
In our experiments, we use a relatively simple model with parameter sharing inspired by NADE, an \gls{ar} neural network designed for density estimation \citep{larochelle2011neural}. More advanced properties like the permutation invariance of MADE \citep{papamakarios2017masked} create an additional overhead that cannot be used in the copula formulation as the predictive update is not permutation-invariant. We refer to Bayesian predictive densities estimated using \gls{ar} neural networks as \textit{ARnet Bayesian predictives} (ARnet-BP). %
\vspace{-2mm}

\paragraph{Tuning the bandwidth function} 
Recall that the bandwidths $\rho_i(\cdot, \cdot)$ are parameterised by $\rho_0$ and the parameters of the chosen covariance functions or neural embedders.
For AR-BP, these are the length scales $\ell$ of the \gls{rbf} covariance function, while for ARnet-BP, these are the parameters $w$ of the \gls{ar} neural network.
We fit these tunable parameters in a data-driven approach by maximising the prequential \citep{dawid1997prequential} log-likelihood $\sum_{i=1}^{n} \log p_{i-1}(x_{i})$ which is analogous to the Bayesian marginal likelihood  -- the tractable predictive density allows us to compute this exactly, and this approach is analogous to empirical Bayes.
Specifically, we use gradient descent optimisation with Adam%
, sampling a different random permutation of the training data at each optimisation step (Supplement \ref{app:ordering}). \vspace{-1mm}

\section{Related Work} 
Our work falls into the broad area of multivariate density estimation \citep{scott2015multivariate}. While \gls{ar} networks have been previously used directly for the task of density estimation \citep{
bengio1999modeling, 
germain2015made, larochelle2011neural}, we use them to elicit a data-dependent bandwidth in the predictive update to mitigate the smoothing effect observed in AR-BP. 
Neural network based approaches, however, often underperform in small-data regimes.
Deep learning approaches that do target few-shot density estimation require complex meta-learning and pre-training pipelines \citep{gu2020ensemble,reed2017few}. %

Our work directly extends the contributions of \citet{hahn2018recursive} and \citet{fong2021martingale} through an alternative specification of the nonparametric Bayesian model in the recursive predictive update scheme. R-BP has recently been used for nonparametric solvency risk prediction \citep{hong2019real}, and survival analysis \citep{fong2022predictive}. \citet{berti2021bayesian, berti2021class, berti2004limit} also focus on univariate predictive updates in the Bayesian nonparametric paradigm, specifically exploring the use of the conditionally identically distributed condition as a relaxation of the standard exchangeability assumption. Other studies have investigated quasi-Bayesian updates in the special case of the mixing distribution in nonparametric mixture models \citep{dixit2022prticle, fortini2020quasi, martin2018nonparametric, tokdar2009consistency}, though these typically focus on univariate or low-dimensional spaces.  See also \cite{martin2021survey} for a survey. 

Finally, copulas are a well-studied tool for modelling the correlations in multivariate data (see e.g. \cite{kauermann2013flexible, ling2020deep, nelsen2007introduction}). Copula density estimation aims to construct density estimates whose univariate marginals are uniform \citep{gijbels1990estimating}, and often focus on modelling strong tail dependencies \citep{wiese2019copula}. In contrast,
we employ bivariate copulas for generic multivariate density estimation as a tool to model the correlations between subsequent subjective predictive densities, rather than across the data dimensions directly. %

\begin{table*}[ht]
\caption{Average NLL with standard error over five runs on data sets analysed by \citet{fong2021martingale}.}
\scriptsize
\begin{tabularx}{\textwidth}{p{0.18\textwidth}YYYYY}
\toprule
{} &                    WINE &                   BREAST &               PARKIN &               IONO &                          BOSTON \\
{n/d} &                89/12 &                   97/14 &               97/16 &               175/30 & 506/13  \\\hline
KDE         &     ${13.69_{\pm 0.00}}$ &    ${10.45_{\pm 0.24}}$ &    ${12.83_{\pm 0.27}}$ &    ${32.06_{\pm 0.00}}$ &      ${8.34_{\pm 0.00}}$ \\
DPMM (Diag) &      ${17.46_{\pm 0.6}}$ &     ${16.26_{\pm 0.71}}$ &     ${22.28_{\pm 0.66}}$ &  ${35.30_{\pm 1.28}}$  &      ${7.64_{\pm 0.09}}$  \\
DPMM (Full) &      ${32.88_{\pm 0.82}}$ &     ${26.67_{\pm 1.32}}$ &     ${39.95_{\pm 1.56}}$ &   ${86.18_{\pm 10.22}}$  &      ${9.45_{\pm 0.43}}$  \\
MAF         &     ${39.60_{\pm 1.41}}$ &    ${10.13_{\pm 0.40}}$ &    ${11.76_{\pm 0.45}}$ &   ${140.09_{\pm 4.03}}$ &    ${56.01_{\pm 27.74}}$ \\
RQ-NSF      &     ${38.34_{\pm 0.63}}$ &    ${26.41_{\pm 0.57}}$ &    ${31.26_{\pm 0.31}}$ &    ${54.49_{\pm 0.65}}$  &     ${-2.20_{\pm 0.11}}$  \\
R-BP        &     ${13.57_{\pm 0.04}}$ &     ${7.45_{\pm 0.02}}$ &     ${9.15_{\pm 0.04}}$ &    ${21.15_{\pm 0.04}}$ &      ${4.56_{\pm 0.04}}$\\
R$_d$-BP    &     ${13.32_{\pm 0.01}}$ &  ${6.12_{\pm 0.05}}$ &     ${7.52_{\pm 0.05}}$ &    ${19.82_{\pm 0.08}}$ &  ${-13.50_{\pm 0.59}}$\\ 
\hdashline[.4pt/4pt]
AR-BP       &     ${13.45_{\pm 0.05}}$ & ${6.18_{\pm 0.05}}$ &     ${8.29_{\pm 0.11}}$ &    ${17.16_{\pm 0.25}}$& ${-0.45_{\pm 0.77}}$ \\ 
AR$_d$-BP   &  $\bm{13.22_{\pm 0.04}}$ &   $\bm{6.11_{\pm 0.04}}$&  $\bm{7.21_{\pm 0.12}}$ &    ${16.48_{\pm 0.26}}$ & $\bm{-14.75_{\pm 0.89}}$\\
ARnet-BP    &      ${14.41_{\pm 0.11}}$ &      ${6.87_{\pm 0.23}}$ &      ${8.29_{\pm 0.17}}$ &  $\bm{15.32_{\pm 0.35}}$ &  ${-5.71_{\pm 0.62}}$\\
\bottomrule
\end{tabularx}
    \label{tab:smalluci}
\end{table*}
\begin{figure*}
    \centering
    \includegraphics[width=0.9\textwidth]{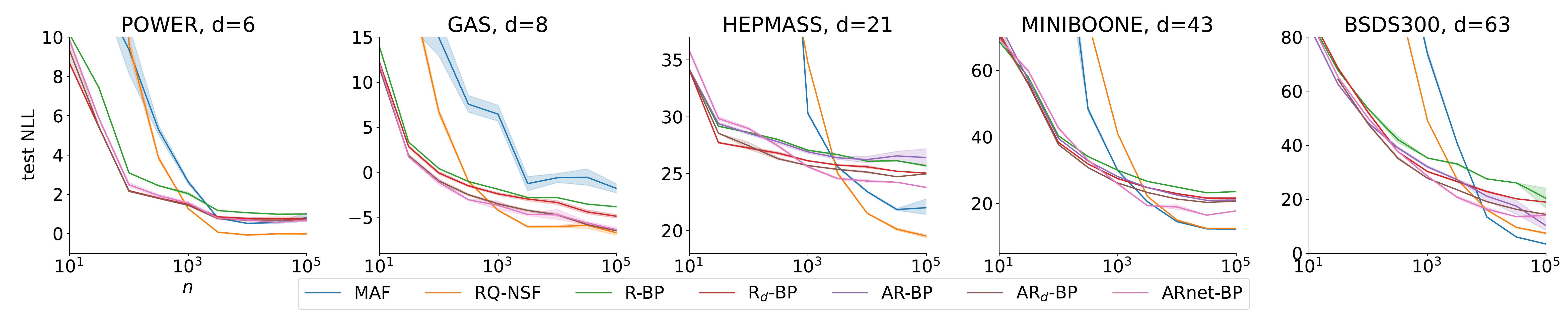}    
    \caption{Average NLL and standard errors over 10 runs for training sets of different size. Our models outperform neural methods for data sets up to 10,000 samples.}
    \label{fig:largeuci} %
\end{figure*}

\section{Experiments} \label{sec:exp}
We demonstrate the benefits of AR-BP, AR$_d$-BP and ARnet-BP for density estimation and prediction tasks in an experimental study with five baseline approaches and 13 different data sets. 
The code and data used is provided in the Supplementary Material. See Supplement \ref{app:exp} for additional experimental details and results, including a sensitivity study, an ablation study, further illustrative examples, a preliminary investigation into image examples, and an empirical study of the computational complexity of the proposed methods. %

\subsection{Density Estimation}
We compared our models against \glspl{kde} \citep{parzen1962estimation}, \glspl{dpmm} \citep{rasmussen1999infinite}, \glspl{maf} \citep{papamakarios2017masked} and \glspl{rqnsf} \citep{durkan2019neural}. 
The hyperparameters of the baselines were tuned with cross-validation. Unless otherwise specified, we use respectively 10 permutations over samples and features to average the quasi-Bayesian estimates. We did not see substantial improvements with more permutations. We use the same few hyperparameters (initialisation of $rho, l_1,\ldots, l_d$, number of permutations, neural network architecture, and learning rate) on all data sets as our method is robust to their choice. 
See Supplement \ref{app:expdetails} for further information.

\paragraph{Data sets analysed by \citet{fong2021martingale}} See Table \ref{tab:smalluci} for the \gls{nll} estimated on five UCI data sets \citep{asuncion2007uci} of small size with up to 506 samples, as investigated by \citet{fong2021martingale}. Our proposed methods display highly competitive performance: AR$_d$-BP achieved the best test \gls{nll} on four of the data sets, while ARnet-BP prevailed on ionosphere.

\begin{table*}[ht]
\scriptsize
\centering
\caption{Average \gls{nll} over five runs reported with standard error for supervised tasks}
\begin{tabularx}{\textwidth}{p{0.12\textwidth}YYYp{0.0001\textwidth}YYY}
\toprule
& \multicolumn{3}{c}{Regression} & & \multicolumn{3}{c}{Classification}\\
{} & BOSTON & CONCR & DIAB & & IONO & PARKIN & MNIST01\\
{n/d} & 506/13 & 1,030/8 & 442/10 & & 351/33 & 195/22 & 12,031/784\\
\cline{2-4} \cline{6-8}
Linear    &     ${0.87_{\pm 0.03}}$ &      ${0.99_{\pm 0.01}}$ &     ${1.07_{\pm 0.01}}$ & &    ${0.33_{\pm 0.01}}$ &     ${0.38_{\pm 0.01}}$ &  ${0.003_{\pm 0.000}}$ \\
GP        &     ${0.42_{\pm 0.08}}$ &      ${0.36_{\pm 0.02}}$ &     ${1.06_{\pm 0.02}}$ & &    ${0.30_{\pm 0.02}}$ &     ${0.42_{\pm 0.02}}$ &      $0.035_{\pm 0.000}$ \\
MLP       &    ${1.42_{\pm 1.01}}$   &     ${2.01_{\pm 0.98}}$ &     ${3.32_{\pm 4.05}}$ & &   ${0.26_{\pm 0.05}}$ &     ${0.31_{\pm 0.02}}$ &     $\bm{0.003_{\pm 0.000}}$ \\ 
R-BP      &     ${0.76_{\pm 0.09}}$ &      ${0.87_{\pm 0.03}}$ &     ${1.05_{\pm 0.03}}$ & &    ${0.26_{\pm 0.01}}$ &     ${0.37_{\pm 0.01}}$ &     ${0.015_{\pm 0.001}}$ \\
R$_d$-BP  &     ${0.40_{\pm 0.03}}$ &      ${0.42_{\pm 0.00}}$ &     ${1.00_{\pm 0.02}}$ & &    ${0.34_{\pm 0.02}}$ &     ${0.27_{\pm 0.03}}$ &     ${0.018_{\pm 0.001}}$ \\
\hdashline[.4pt/4pt]
AR-BP     &     ${0.52_{\pm 0.13}}$ &      ${0.42_{\pm 0.01}}$ &     ${1.06_{\pm 0.02}}$ & &    ${0.21_{\pm 0.02}}$ &     ${0.29_{\pm 0.02}}$ &    ${0.015_{\pm 0.001}}$ \\
AR$_d$-BP &  $\bm{0.37_{\pm 0.10}}$ &      ${0.39_{\pm 0.01}}$ &  $\bm{0.99_{\pm 0.02}}$ & & $\bm{0.20_{\pm 0.02}}$ &     ${0.28_{\pm 0.03}}$ &    ${0.017_{\pm 0.001}}$ \\
ARnet-BP  &     ${0.45_{\pm 0.11}}$ &  $\bm{-0.03_{\pm 0.00}}$ &     ${1.41_{\pm 0.07}}$ & &    ${0.24_{\pm 0.04}}$ &  $\bm{0.26_{\pm 0.04}}$ &    ${0.014_{\pm 0.001}}$ \\
\bottomrule
\end{tabularx}
    \label{tab:preduci}
\end{table*}

\paragraph{Data sets analysed by \citet{papamakarios2017masked}} A number of UCI data sets have become the standard evaluation benchmark for deep \gls{ar} models \citep{durkan2019neural, huang2018neural, papamakarios2017masked}. These include low-dimensional data sets with up to 63 features, but at least 29,000 with up to $10^{6}$ samples. In many circumstances, data sets of such a data size are not avialable. To investigate performance as a function of sample size, we trained the models on subsets of the full data set. We do not report results for the \glspl{kde} and the \gls{dpmm} estimators here as these estimators performed significantly worse than the other approaches. Similarly, we do not report deep learning results for sample sizes smaller than $10^2$. See Supplement \ref{app:otherexp} for complete results. 

In the small-data regime, we observe that the R-BP methods significantly outperform the neural density estimators (Figure \ref{fig:largeuci}). As the sample size increases, the gap in performance decreases until eventually the neural density estimators outcompete the R-BP methods. The performance between the R-BP methods and our proposed \gls{ar} extensions is largely similar, though we note that the AR-BP methods were generally more effective on the GAS dataset.

\subsection{Supervised Learning}
R-BP methods, including AR-BP, can be used for prediction tasks such as regression and classification \cite{fong2021martingale}. In short, this is achieved by estimating the conditional predictive density $p_n(y|x)$ of the labels $y$ directly by assuming a dependent Dirichlet process likelihood. See Supplement \ref{app:supervised} for details. Again, we follow the experimental set-up of \citet{fong2021martingale}, and additionally report results on the MNIST data set, restricted to digits of class 0 and 1. We report the conditional test \gls{nll} $-\frac{1}{\ns}\sum_{i}%
\log p_n(y^*_i|x^*_i)$ for a test set $\{(x^*_1, y^*_1), \ldots, (x^*_{\ns}, y^*_{\ns})\}$. 
We compared our models against a \gls{gp}, a linear Bayesian model (Linear), and a one-hidden-layer multilayer perceptron (MLP) on several classification and regression tasks. To get a distribution over the predicted outcome in the regression case, we trained an ensemble over 10 MLPs. Our proposed methods were again highly competitive (Table \ref{tab:preduci}). AR$_d$-BP performed best on two regression tasks and one classification task. ARnet-BP was substantially better than the remaining methods on CONCR and also performed best on the PARKIN. On the other hand, the MLP model was best on MNIST.

\section{Discussion} \label{sec:concl}
Although Bayesian methods generally perform well in the small sample setting, the conventional Bayesian approach to density estimation, i.e. \gls{dpmm} estimation via the posterior predictive, is computationally intensive. Here, we set out to propose a computationally efficient density estimator as an alternative to \gls{dpmm} density estimation. We recommend its use for tabular data sets of up to 63 features, and 10,000 observations. Such data set sizes are ubiquitous in healthcare, finance, hyperparameter tuning, and survey data applications.

We expand upon the tractable recursive copula updates of \citet{fong2021martingale, hahn2018recursive} by incorporating regression methods, such as kernels and neural networks. This introduces a data-dependent bandwidth, thus increasing the flexibility of this class of models, with little computational overhead compared to R-BP. More generally, it would be of interest to integrate other machine learning methods with recursive copula updates. Furthermore, other Bayesian nonparametric models may inspire other recursive copula updates -- see Appendix \ref{app:deriv-gp} for an example based on \glspl{gp}. 

An appealing feature of AR-BP is that it requires no manual hyperparameter tuning. Further, on small data sets, AR-BP shows state-of-the-art generalization and is faster than competing deep learning models. It significantly increases the modelling capacity of the baseline R-BP via a data-dependent bandwidth. Additionally, ARnet-BP provides a useful illustration of how powerful neural network models can be incorporated into R-BP methods to improve density estimation. Future work can investigate alternative architectures for structured data. Our work adds to the rich body of density estimators and thus we do not anticipate any additional negative societal impact arising from our proposal. 

This strong performance of AR-BP (and other copula methods) in the small data regime is likely due to its Bayesian-like regularization towards an initial density $p_0$, as shown in the weighted sum in \eqref{eq:mv_DP_AR_marginal}. Its weaker performance in the large data regime may be due to the importance of the sequence $\alpha_i$ which governs how regularization decays, but further theoretical work is needed to understand AR-BP's asymptotic behaviour.
A limitation of R-BP methods, including AR-BP, is the quadratic time dependence on the number of training observations. Subsampling techniques thus offer a particularly promising avenue to reduce the overall computational cost and warrant further investigation. Although the recursive updates depend on the sample and covariate ordering, it is possible to alleviate this dependence though by estimating the R-BP over multiple permutations in parallel, as we have done in the above experiments. Nevertheless, the algorithm is relatively fast: with a single GPU, we were able to train models with 100,000 observations in less than an hour.

The use of a \gls{gp} prior greatly increases the flexibility of our framework. Moreover, it opens the door to future research to incorporate ideas from the vast \gls{gp} literature to further boost performance in high-dimensional settings. Our use of the \gls{rbf} kernel was illustrative; other kernels are discussed in Appendix \ref{app:exp-add} where we find that the \gls{rbf} kernel is . For example, we anticipate that the use of recent advances in convolutional kernels \citep{van2017convolutional} would be particularly suited for computer vision tasks.

\clearpage
\bibliographystyle{plainnat}
\bibliography{bib}

\begin{thebibliography}{51}
\providecommand{\natexlab}[1]{#1}
\providecommand{\url}[1]{\texttt{#1}}
\expandafter\ifx\csname urlstyle\endcsname\relax
  \providecommand{\doi}[1]{doi: #1}\else
  \providecommand{\doi}{doi: \begingroup \urlstyle{rm}\Url}\fi

\bibitem[Asuncion and Newman(2007)]{asuncion2007uci}
Arthur Asuncion and David Newman.
\newblock {UCI} machine learning repository, 2007.

\bibitem[Bengio and Bengio(1999)]{bengio1999modeling}
Yoshua Bengio and Samy Bengio.
\newblock Modeling high-dimensional discrete data with multi-layer neural
  networks.
\newblock \emph{Advances in Neural Information Processing Systems}, 12, 1999.

\bibitem[Berti et~al.(2004)Berti, Pratelli, and Rigo]{berti2004limit}
Patrizia Berti, Luca Pratelli, and Pietro Rigo.
\newblock Limit theorems for a class of identically distributed random
  variables.
\newblock \emph{The Annals of Probability}, 32\penalty0 (3):\penalty0
  2029--2052, 2004.

\bibitem[Berti et~al.(2021{\natexlab{a}})Berti, Dreassi, Leisen, Rigo, and
  Pratelli]{berti2021bayesian}
Patrizia Berti, Emanuela Dreassi, Fabrizio Leisen, Pietro Rigo, and Luca
  Pratelli.
\newblock Bayesian predictive inference without a prior.
\newblock \emph{arXiv preprint arXiv:2104.11643}, 2021{\natexlab{a}}.

\bibitem[Berti et~al.(2021{\natexlab{b}})Berti, Dreassi, Pratelli, and
  Rigo]{berti2021class}
Patrizia Berti, Emanuela Dreassi, Luca Pratelli, and Pietro Rigo.
\newblock A class of models for {B}ayesian predictive inference.
\newblock \emph{Bernoulli}, 27\penalty0 (1):\penalty0 702--726,
  2021{\natexlab{b}}.

\bibitem[Chopin(2002)]{chopin2002sequential}
Nicolas Chopin.
\newblock A sequential particle filter method for static models.
\newblock \emph{Biometrika}, 89\penalty0 (3):\penalty0 539--552, 2002.

\bibitem[Dawid(1997)]{dawid1997prequential}
A~Philip Dawid.
\newblock Prequential analysis.
\newblock \emph{Encyclopedia of Statistical Sciences}, 1:\penalty0 464--470,
  1997.

\bibitem[De~Finetti(1937)]{de1937prevision}
Bruno De~Finetti.
\newblock La pr{\'e}vision: ses lois logiques, ses sources subjectives.
\newblock \emph{Annales de l'institut Henri Poincar{\'e}}, 7:\penalty0 1--68,
  1937.

\bibitem[Dinh et~al.(2014)Dinh, Krueger, and Bengio]{dinh2014nice}
Laurent Dinh, David Krueger, and Yoshua Bengio.
\newblock Nice: Non-linear independent components estimation.
\newblock \emph{arXiv preprint arXiv:1410.8516}, 2014.

\bibitem[Dixit and Martin(2022)]{dixit2022prticle}
Vaidehi Dixit and Ryan Martin.
\newblock A prticle filter algorithm for nonparametric estimation of
  multivariate mixing distributions.
\newblock \emph{arXiv preprint arXiv:2204.01646}, 2022.

\bibitem[Durkan et~al.(2019)Durkan, Bekasov, Murray, and
  Papamakarios]{durkan2019neural}
Conor Durkan, Artur Bekasov, Iain Murray, and George Papamakarios.
\newblock Neural spline flows.
\newblock \emph{Advances in neural information processing systems}, 32, 2019.

\bibitem[Escobar and West(1995)]{escobar1995bayesian}
Michael~D Escobar and Mike West.
\newblock Bayesian density estimation and inference using mixtures.
\newblock \emph{Journal of the american statistical association}, 90\penalty0
  (430):\penalty0 577--588, 1995.

\bibitem[Escobar(1988)]{escobar1988estimating}
Michael~David Escobar.
\newblock \emph{Estimating the means of several normal populations by
  nonparametric estimation of the distribution of the means}.
\newblock PhD thesis, Yale University, 1988.

\bibitem[Ferguson(1973)]{ferguson1973bayesian}
Thomas~S Ferguson.
\newblock A {B}ayesian analysis of some nonparametric problems.
\newblock \emph{The annals of statistics}, pages 209--230, 1973.

\bibitem[Fong and Lehmann(2022)]{fong2022predictive}
Edwin Fong and Brieuc Lehmann.
\newblock A predictive approach to bayesian nonparametric survival analysis.
\newblock In Gustau Camps-Valls, Francisco J.~R. Ruiz, and Isabel Valera,
  editors, \emph{Proceedings of The 25th International Conference on Artificial
  Intelligence and Statistics}, volume 151 of \emph{Proceedings of Machine
  Learning Research}, pages 6990--7013. PMLR, 28--30 Mar 2022.
\newblock URL \url{https://proceedings.mlr.press/v151/fong22a.html}.

\bibitem[Fong et~al.(2021)Fong, Holmes, and Walker]{fong2021martingale}
Edwin Fong, Chris Holmes, and Stephen~G Walker.
\newblock Martingale posterior distributions.
\newblock \emph{To appear at the Journal of the Royal Statistical Society:
  Series B (with discussion)}, 2021.

\bibitem[Fortini and Petrone(2020)]{fortini2020quasi}
Sandra Fortini and Sonia Petrone.
\newblock Quasi-bayes properties of a procedure for sequential learning in
  mixture models.
\newblock \emph{Journal of the Royal Statistical Society: Series B (Statistical
  Methodology)}, 82\penalty0 (4):\penalty0 1087--1114, 2020.

\bibitem[Frey et~al.(1998)Frey, Brendan, and Frey]{frey1998graphical}
Brendan~J Frey, J~Frey Brendan, and Brendan~J Frey.
\newblock \emph{Graphical models for machine learning and digital
  communication}.
\newblock MIT press, 1998.

\bibitem[Germain et~al.(2015)Germain, Gregor, Murray, and
  Larochelle]{germain2015made}
Mathieu Germain, Karol Gregor, Iain Murray, and Hugo Larochelle.
\newblock Made: Masked autoencoder for distribution estimation.
\newblock In \emph{International Conference on Machine Learning}, pages
  881--889. PMLR, 2015.

\bibitem[Gijbels and Mielniczuk(1990)]{gijbels1990estimating}
Iène Gijbels and Jan Mielniczuk.
\newblock Estimating the density of a copula function.
\newblock \emph{Communications in Statistics - Theory and Methods}, 19\penalty0
  (2):\penalty0 445--464, January 1990.
\newblock ISSN 0361-0926.
\newblock \doi{10.1080/03610929008830212}.
\newblock URL \url{https://doi.org/10.1080/03610929008830212}.

\bibitem[Gu et~al.(2020)Gu, Zhang, and Qiao]{gu2020ensemble}
Ke~Gu, Yonghui Zhang, and Junfei Qiao.
\newblock Ensemble meta-learning for few-shot soot density recognition.
\newblock \emph{IEEE Transactions on Industrial Informatics}, 17\penalty0
  (3):\penalty0 2261--2270, 2020.

\bibitem[Gunawan et~al.(2020)Gunawan, Dang, Quiroz, Kohn, and
  Tran]{gunawan2020subsampling}
David Gunawan, Khue-Dung Dang, Matias Quiroz, Robert Kohn, and Minh-Ngoc Tran.
\newblock Subsampling sequential monte carlo for static bayesian models.
\newblock \emph{Statistics and Computing}, 30\penalty0 (6):\penalty0
  1741--1758, 2020.

\bibitem[Hahn et~al.(2018)Hahn, Martin, and Walker]{hahn2018recursive}
P~Richard Hahn, Ryan Martin, and Stephen~G Walker.
\newblock On recursive {B}ayesian predictive distributions.
\newblock \emph{Journal of the American Statistical Association}, 113\penalty0
  (523):\penalty0 1085--1093, 2018.

\bibitem[Hennigan et~al.(2020)Hennigan, Cai, Norman, and
  Babuschkin]{haiku2020github}
Tom Hennigan, Trevor Cai, Tamara Norman, and Igor Babuschkin.
\newblock {H}aiku: {S}onnet for {JAX}, 2020.
\newblock URL \url{http://github.com/deepmind/dm-haiku}.

\bibitem[Hewitt and Savage(1955)]{hewitt1955symmetric}
Edwin Hewitt and Leonard~J Savage.
\newblock Symmetric measures on cartesian products.
\newblock \emph{Transactions of the American Mathematical Society}, 80\penalty0
  (2):\penalty0 470--501, 1955.

\bibitem[Hill(1968)]{hill1968posterior}
Bruce~M Hill.
\newblock Posterior distribution of percentiles: Bayes' theorem for sampling
  from a population.
\newblock \emph{Journal of the American Statistical Association}, 63\penalty0
  (322):\penalty0 677--691, 1968.

\bibitem[Hong and Martin(2019)]{hong2019real}
Liang Hong and Ryan Martin.
\newblock Real-time {B}ayesian non-parametric prediction of solvency risk.
\newblock \emph{Annals of Actuarial Science}, 13\penalty0 (1):\penalty0 67--79,
  2019.

\bibitem[Huang et~al.(2018)Huang, Krueger, Lacoste, and
  Courville]{huang2018neural}
Chin-Wei Huang, David Krueger, Alexandre Lacoste, and Aaron Courville.
\newblock Neural autoregressive flows.
\newblock In \emph{International Conference on Machine Learning}, pages
  2078--2087. PMLR, 2018.

\bibitem[Kauermann et~al.(2013)Kauermann, Schellhase, and
  Ruppert]{kauermann2013flexible}
G{\"o}ran Kauermann, Christian Schellhase, and David Ruppert.
\newblock Flexible copula density estimation with penalized hierarchical
  b-splines.
\newblock \emph{Scandinavian Journal of Statistics}, 40\penalty0 (4):\penalty0
  685--705, 2013.

\bibitem[Kingma et~al.(2016)Kingma, Salimans, Jozefowicz, Chen, Sutskever, and
  Welling]{kingma2016improved}
Durk~P Kingma, Tim Salimans, Rafal Jozefowicz, Xi~Chen, Ilya Sutskever, and Max
  Welling.
\newblock Improved variational inference with inverse autoregressive flow.
\newblock \emph{Advances in neural information processing systems}, 29, 2016.

\bibitem[Larochelle and Murray(2011)]{larochelle2011neural}
Hugo Larochelle and Iain Murray.
\newblock The neural autoregressive distribution estimator.
\newblock In \emph{Proceedings of the fourteenth international conference on
  artificial intelligence and statistics}, pages 29--37. JMLR Workshop and
  Conference Proceedings, 2011.

\bibitem[Ling et~al.(2020)Ling, Fang, and Kolter]{ling2020deep}
Chun~Kai Ling, Fei Fang, and J~Zico Kolter.
\newblock Deep archimedean copulas.
\newblock \emph{Advances in Neural Information Processing Systems},
  33:\penalty0 1535--1545, 2020.

\bibitem[Lueckmann et~al.(2021)Lueckmann, Boelts, Greenberg, Goncalves, and
  Macke]{lueckmann2021benchmarking}
Jan-Matthis Lueckmann, Jan Boelts, David Greenberg, Pedro Goncalves, and Jakob
  Macke.
\newblock Benchmarking simulation-based inference.
\newblock In \emph{International Conference on Artificial Intelligence and
  Statistics}, pages 343--351. PMLR, 2021.

\bibitem[Martin(2018)]{martin2018nonparametric}
Ryan Martin.
\newblock On nonparametric estimation of a mixing density via the predictive
  recursion algorithm.
\newblock \emph{arXiv preprint arXiv:1812.02149}, 2018.

\bibitem[Martin(2021)]{martin2021survey}
Ryan Martin.
\newblock A survey of nonparametric mixing density estimation via the
  predictive recursion algorithm.
\newblock \emph{Sankhya B}, 83\penalty0 (1):\penalty0 97--121, 2021.

\bibitem[Nelsen(2007)]{nelsen2007introduction}
R.B. Nelsen.
\newblock \emph{An {Introduction} to {Copulas}}.
\newblock Springer {Series} in {Statistics}. Springer New York, 2007.
\newblock ISBN 978-0-387-28678-5.

\bibitem[Papamakarios et~al.(2017)Papamakarios, Pavlakou, and
  Murray]{papamakarios2017masked}
George Papamakarios, Theo Pavlakou, and Iain Murray.
\newblock Masked autoregressive flow for density estimation.
\newblock \emph{Advances in neural information processing systems}, 30, 2017.

\bibitem[Parzen(1962)]{parzen1962estimation}
Emanuel Parzen.
\newblock On estimation of a probability density function and mode.
\newblock \emph{The annals of mathematical statistics}, 33\penalty0
  (3):\penalty0 1065--1076, 1962.

\bibitem[Rasmussen(1999)]{rasmussen1999infinite}
Carl Rasmussen.
\newblock The infinite gaussian mixture model.
\newblock \emph{Advances in neural information processing systems}, 12, 1999.

\bibitem[Reed et~al.(2017)Reed, Chen, Paine, Oord, Eslami, Rezende, Vinyals,
  and de~Freitas]{reed2017few}
Scott Reed, Yutian Chen, Thomas Paine, A{\"a}ron van~den Oord, SM~Eslami,
  Danilo Rezende, Oriol Vinyals, and Nando de~Freitas.
\newblock Few-shot autoregressive density estimation: Towards learning to learn
  distributions.
\newblock \emph{arXiv preprint arXiv:1710.10304}, 2017.

\bibitem[Salakhutdinov and Murray(2008)]{salakhutdinov2008quantitative}
Ruslan Salakhutdinov and Iain Murray.
\newblock On the quantitative analysis of deep belief networks.
\newblock In \emph{Proceedings of the 25th international conference on Machine
  learning}, pages 872--879, 2008.

\bibitem[Scaldelai et~al.(2022)Scaldelai, Matioli, Santos, and
  Kleina]{scaldelai2022multiclusterkde}
D~Scaldelai, LC~Matioli, SR~Santos, and M~Kleina.
\newblock Multiclusterkde: a new algorithm for clustering based on multivariate
  kernel density estimation.
\newblock \emph{Journal of Applied Statistics}, 49\penalty0 (1):\penalty0
  98--121, 2022.

\bibitem[Scott(2015)]{scott2015multivariate}
David~W Scott.
\newblock \emph{Multivariate density estimation: theory, practice, and
  visualization}.
\newblock John Wiley \& Sons, 2015.

\bibitem[Sklar(1959)]{sklar1959fonctions}
M~Sklar.
\newblock Fonctions de repartition an dimensions et leurs marges.
\newblock \emph{Publ. inst. statist. univ. Paris}, 8:\penalty0 229--231, 1959.

\bibitem[Tokdar et~al.(2009)Tokdar, Martin, and Ghosh]{tokdar2009consistency}
Surya~T Tokdar, Ryan Martin, and Jayanta~K Ghosh.
\newblock Consistency of a recursive estimate of mixing distributions.
\newblock \emph{The Annals of Statistics}, pages 2502--2522, 2009.

\bibitem[Van~der Wilk et~al.(2017)Van~der Wilk, Rasmussen, and
  Hensman]{van2017convolutional}
Mark Van~der Wilk, Carl~Edward Rasmussen, and James Hensman.
\newblock Convolutional gaussian processes.
\newblock \emph{Advances in Neural Information Processing Systems}, 30, 2017.

\bibitem[Vinyals et~al.(2015)Vinyals, Bengio, and Kudlur]{vinyals2015order}
Oriol Vinyals, Samy Bengio, and Manjunath Kudlur.
\newblock Order matters: Sequence to sequence for sets.
\newblock \emph{arXiv preprint arXiv:1511.06391}, 2015.

\bibitem[Wade et~al.(2014)Wade, Walker, and Petrone]{wade2014predictive}
Sara Wade, Stephen~G Walker, and Sonia Petrone.
\newblock A predictive study of dirichlet process mixture models for curve
  fitting.
\newblock \emph{Scandinavian Journal of Statistics}, 41\penalty0 (3):\penalty0
  580--605, 2014.

\bibitem[Wiese et~al.(2019)Wiese, Knobloch, and Korn]{wiese2019copula}
Magnus Wiese, Robert Knobloch, and Ralf Korn.
\newblock Copula \& marginal flows: Disentangling the marginal from its joint.
\newblock \emph{arXiv preprint arXiv:1907.03361}, 2019.

\bibitem[Williams and Rasmussen(2006)]{williams2006gaussian}
Christopher~K Williams and Carl~Edward Rasmussen.
\newblock \emph{Gaussian processes for machine learning}.
\newblock Number~3 in 2. MIT press Cambridge, MA, 2006.

\bibitem[Zoran and Weiss(2011)]{zoran2011learning}
Daniel Zoran and Yair Weiss.
\newblock From learning models of natural image patches to whole image
  restoration.
\newblock In \emph{2011 International Conference on Computer Vision}, pages
  479--486. IEEE, 2011.

\end{thebibliography}

\clearpage
\appendix
\onecolumn
\section{Derivations} \label{app:deriv}

\subsection{Derivation of AR-BP}  \label{app:ar-deriv}
For illustration purposes, we first start by summarising the derivation of the update without autoregression, closely following Appendix E.1.2 in \citet{fong2021martingale}.

\subsubsection{No Autoregression (R-BP)} \label{Appendix:multivariate}
The multivariate DPMM with factorized kernel has the form
\begin{equation*}
\begin{aligned}
f_G({x}) = \int \prod_{j=1}^d\mathcal{N}(x^j \mid \theta^j,1) \, dG({\theta}),\quad 
 G \sim \text{DP}\left(a, G_0 \right), \quad G_0({\theta}) = \prod_{j=1}^d\mathcal{N}(\theta^j \mid 0,\tau^{-1}).
\end{aligned}
\end{equation*}
Given 
\begin{align*}
    p_i(x) = p_{i-1}(x)h_i(x, x_i),
\end{align*}
\citet{hahn2018recursive} and \citet{fong2021martingale} derive the predictive density updates for R-BP by initally only considering the first step update $h_1$
\begin{align*}p_{1}(x) = p_{0}(x)h_1(x, x_1)\cdot\end{align*}
From 
\begin{align*}
        h_i(x, x_i) = &\frac{\int f(x|\theta)f(x_i|\theta)\pi_{i-1}(\theta) d\theta}{\int f(x|\theta)\pi_{i-1}(\theta)d\theta\int f(x_n|\theta)\pi_{i-1}(\theta)d\theta}, %
\end{align*}
it follows that
\begin{align}\label{eq:mv_num}
h_1(x, x_1) = \frac{E\left[ f_G({x})\, f_G({x}_1) \right]}{p_0({x})\, p_0({x}_1)}
\end{align}
where the expectation is over $G$ coming from the prior. Following the stick-breaking representation of the DP, \citet{fong2021martingale} write $G$ as
\begin{align*}
G = \sum_{k=1}^\infty w_k \, \delta_{{\theta}^*_k}
\end{align*}
where $w_k = v_k \prod_{j< k} \{1-v_j\}$, $v_k \iid \text{Beta}(1,a)$ and ${\theta}^*_k \iid G_0$.
\citet{fong2021martingale} then derive the numerator as
\begin{align*}
&E\left[ \sum_{j=1}^\infty \sum_{k=1}^\infty  w_j \,w_k\, K({x} \mid {\theta}^*_j)\, K({x}_1 \mid {\theta}^*_k) \right] %
\\&=\left(1-E\left[ \sum_{k=1}^\infty w_k^2\right]\right)E\left[ K({x} \mid {\theta}^*) \right] E\left[ K({x}_1 \mid {\theta}^*) \right] + E\left[ \sum_{k=1}^\infty w_k^2\right] E\left[ K({x} \mid {\theta}^*)\, K({x}_1 \mid {\theta}^*) \right] 
\end{align*}
where they have used the fact that $\sum_{k=1}^\infty w_k = 1$ almost surely. 
As $p_0({x}) = E\left[ K({x} \mid {\theta}^*) \right]$, it follows that \eqref{eq:mv_num} can be expressed as
\begin{align*}
1-\alpha_1 + \alpha_1 \frac{E\left[ K({x} \mid {\theta}^*) \, K({x}_1 \mid {\theta}^*) \right] }{p_0({x}) \, p_0({x}_1)} \cdot
\end{align*}
for some fixed $\alpha_1$.
For R-BP, the kernel $K$ factorises with independent priors on each dimension, and $p_0({x}) = \prod_{j=1}^d p_0(x^j) = \prod_{j=1}^d \mathcal{N}(x^j \mid 0,1 +\tau^{-1})$, so
\begin{align}
\frac{E\left[ K({x} \mid {\theta}^*) \, K({x}_1 \mid {\theta}^*) \right] }{p_0({x})\, p_0({x}_1)}  = \prod_{j=1}^d \frac{E\left[ K(x^j \mid \theta^{*j}) \, K({x}^j_1 \mid {\theta^{*j}}) \right] }{p_0({x}^j) \, p_0({x}^j_1)}\cdot \label{eq:jointexp}
\end{align}
\citet{fong2021martingale} then show that each univariate term corresponds to the bivariate Gaussian copula density, \begin{align*}c(u,v; \rho) =  \frac{\mathcal{N}_2\left\{\Phi^{-1}(u),\Phi^{-1}(v) \mid 0,1,\rho \right\} }{\mathcal{N}\left\{\Phi^{-1}(u)\mid 0,1\right\}\mathcal{N}\left\{\Phi^{-1}(v)\mid 0,1\right\}},\end{align*}

where $\Phi$ is the normal \gls{cdf}, and $\mathcal{N}_2$ is the standard bivariate density with correlation parameter $\rho = 1/(1+\tau)$. They then suggest an alternative sequence $h_i$ which iteratively repeats $h_1$, with the key feature that  $\alpha_i=(2-\frac{1}{i})\frac{1}{i+1}$. See Appendix E.1.1. in \cite{fong2021martingale} for a derivation of this sequence $\alpha_i$.

\subsubsection{With Autoregression (AR-BP)} 

For the derivation of the AR-BP update, we can follow the arguments in the previous section until \eqref{eq:jointexp} where the factorised kernel assumption applies for the first time. For AR-BP, we instead have
\begin{align}
\frac{E\left[ K({x} \mid {\theta}^*) \, K({x}_1 \mid {\theta}^*) \right] }{p_0({x})\, p_0({x}_1)}  = \prod_{j=1}^d \frac{E\left[ K\{x^j \mid \theta^{*j}(x^{1:j-1}) \}\, K\{{x}^j_1 \mid {\theta^{*j}(x^{1:j-1})\}} \right] }{p_0({x}^j) \, p_0({x}^j_1)}\cdot %
\end{align}
The factorisation of the denominator follows from 
\begin{align*}p_0({x})=E\left[\prod_{j=1}^d K\{x^j \mid \theta^{*j}(x^{1:j-1})\} \right] = \prod_{j=1}^d E\left[K\{x^j \mid \theta^{*j}(x^{1:j-1})\} \right]\end{align*}
as we have independent GP priors on each function $\theta^{*j}$. For notational convenience we write $\{y,x\}$ in place of $\{x^j, x^{1:j-1}\}$ in the following.
With the autoregressive kernel assumption, there is the additional complexity
\begin{align*}
E\left[\mathcal{N}\{y \mid \theta(x), 1\}\, \mathcal{N}\{y_1 \mid \theta(x_1), 1\} \right]
\end{align*}
where $\theta(\cdot) \sim\text{GP}\{0,\tau^{-1}k\}$. The marginal distribution of the GP is normal, so we have
\begin{align*}\left[ \theta(x), \theta(x_1)\right]^\T \sim \mathcal{N}_2(x, x_1 \mid 0, \Sigma_{x,x_1})\end{align*} 
where
\begin{align*}
\Sigma_{x,x_1} = \begin{bmatrix}\tau^{-1} &\tau^{-1} k(x,x_1) \\\tau^{-1} k(x,x_1) & \tau^{-1}
\end{bmatrix}\cdot
\end{align*}
Again from the conjugacy of the normal, we can show that
\begin{align*}
E\left[\mathcal{N}\{y \mid \theta(x), 1\} \mathcal{N}\{y_1 \mid \theta(x_1), 1\}  \right] = \mathcal{N}(y,y_1 \mid 0, K_{x,x_1})
\end{align*}
where 
\begin{align*}
K_{x,x_1} = \begin{bmatrix} 1+\tau^{-1} &\tau^{-1} k(x,x_1) \\\tau^{-1} k(x,x_1) & 1+\tau^{-1}
\end{bmatrix}\cdot
\end{align*}
Here $p_0(y)=E[\mathcal{N}(y|\theta(x))]$ is the same as above, since marginally $\theta(x) \sim \mathcal{N}(0,\tau^{-1})$. Plugging in $y = P_0^{-1}\{\Phi(z)\}$ again gives us the Gaussian copula density with correlation parameter
\begin{align*}
\rho_1(x) = \rho_0 k(x,x_1)
\end{align*}
for $\rho_0 = 1/(1+\tau)$.

\subsection{Derivation of Gaussian Process Posterior} \label{app:deriv-gp}
{In this section, we derive the copula sequence for the Gaussian Process, which is fully tractable. This section is mostly for insight, but it would however be interesting to investigate any potential avenues for methodological development.}
\subsubsection{First Update Step}

{We consider a univariate regression setting with $\{y,x\}$. For the GP, we have the model
\begin{align*}f_\theta(y \mid x) = \mathcal{N}(y \mid \theta(x), \sigma^2), \quad \theta(\cdot) \sim \text{GP}(0, \tau^{-1} k).
\end{align*}
Like in the above, we can derive the function $h_1(x,x_1)$.
}
Following a similar argument to the AR-BP derivation, the first step GP copula density is
\begin{equation*}
\begin{aligned}
 \frac{\Normal_2\left(y,y_1\mid 0,K_2+ \sigma^2I \right)}{p_0(y \mid x) p_0(y_1 \mid x_1)}
\end{aligned}
\end{equation*}
where $K_i$ is the $i\times i$ Gram matrix, with kernel
\begin{equation*}
 k(x,x') = \tau^{-1} \exp\left\{ -0.5 (x-x')^2/\ell \right\}. 
 \end{equation*}
Writing in terms of $P_0$, we have
\begin{equation*}
\begin{aligned}
c\left\{P_0(y\mid x), P_0(y_1 \mid x_1); {\rho_1(x)} \right\}
\end{aligned}
\end{equation*}
where $c$ is again the Gaussian copula density, but we have the correlation parameter as 
\begin{equation*}
\rho_1(x)  = \frac{\exp\left\{-0.5(x-x_1)^2/\ell\right\}}{ 1+ \tau\sigma^2}.
\end{equation*}
From this, we can derive the first step of the update scheme:
\begin{align*}
p_1(y \mid x) &=  c\{P_{0}(y\mid x), P_{0}(y_1 \mid x_1); {\rho_1(x)}\} \,  p_{0}(y \mid x)
\end{align*}
where $c(u,v ; \rho)$ is again the Gaussian copula density, and $p_0(y\mid x) = \Normal(y; 0, \sigma^2 + \tau^{-1})$. %

\subsubsection{All Update Steps} \label{app:gp}
We can even derive the copula update scheme for $i> 1$, as the Gaussian process posterior is tractable. After observing $i-1$ observations, we have
\begin{equation*}
\begin{aligned}
\pi(\theta_x, \theta_{x_{i}} \mid y_{1:{i-1}},x_{1:{i-1}}) &= \mathcal{N}(\mu_{i-1}, \Sigma_{i-1})\\
\end{aligned}
\end{equation*}
where each element of $\Sigma_{i-1}$ has the entry
 \begin{equation*}
 k_{i-1}(x,x') = k(x,x')  - k(x, x_{1:i-1})\left[K_{i-1} + \sigma^2 I \right]^{-1}k(x_{1:i-1},x')
 \end{equation*}
 where the subscript $i-1$ indicates it is the posterior kernel and $\mu_{i-1}$ is the posterior mean vector of the GP at $x$ and $x_i$. Marginally, the GP copula after $i-1$ data points is
\begin{equation*}
\begin{aligned}
 \frac{\mathcal{N}_2\left(y,y_{i}; \mu_{i-1},\Sigma_{i-1}+ \sigma^2I \right)}{\mathcal{N}\left\{y; \mu_{i-1}^{y}, k_{i-1}(x,x)+  \sigma^2 \right\} \mathcal{N}\left\{y_{i+1};  \mu_{i-1}^{y_{i-1}}, k_{i-1}(x_{i},x_{i})+ \sigma^2 \right\}}
\end{aligned}
\end{equation*} 
where $\mu_{i-1}^y$ is the posterior mean of the GP at $x$ and likewise for $\mu_{i-1}^{y_{i-1}}$. This is equivalent to the bivariate Gaussian copula density $c(u,v; \rho_i(x)),$
where as before $u = P_{i-1}(y \mid x)$ and $v = P_{i-1}(y_{i+1} \mid x_{i+1})$. The correlation parameter is now
 \begin{equation*} %
\rho_i(x) = \frac{k_{i-1}(x,x_i)}{\sqrt{\{k_{i-1}(x,x) + \sigma^2\}\{k_{i-1}(x_i,x_i) + \sigma^2\}}}
 \end{equation*}
 In summary, we have the update
 \begin{align*}
p_i(y \mid x) &=  c\{P_{i-1}(y\mid x), P_{i-1}(y_i \mid x_i); {\rho_i(x)}\} \,  p_{i-1}(y \mid x).
\end{align*}
This gives the same predictives as fitting a full GP. While this update form does not offer any computational gains, it gives us insight into the GP update. The copula update corresponds to the regular normal update \citep{hahn2018recursive} with a data-dependent bandwidth $\rho_i(x)$ which measures the distance between $x$ and $x_i$ based on the posterior kernel. A potential interesting direction of research is to seek approximations of the expensive $\rho_i(x)$ to aid with the computation of the GP.
\subsection{Intuition for AR Copula} \label{app:cond_mean}

As in the main paper, we consider bivariate data, $(x,y)$. As shown in \cite{fong2021martingale}, the update for the conditional density for R-BP takes the form
\begin{equation}\label{eq:app_condit}
p_i(y \mid x) = [1-\alpha_i(x,x_i) + \alpha_i(x,x_i) \, c\left\{P_{i-1}(y \mid x), P_{i-1}( y_i \mid x_i); \rho \right\}]\, p_{i-1}(y \mid x),
\end{equation}
where
\begin{align*}
\alpha_i(x,x_i) = \frac{\alpha_i c\{P_{i-1}(x), P_{i-1}(x_i); \rho\}}{1-\alpha_i + \alpha_i c\{P_{i-1}(x), P_{i-1}(x_i); \rho\}}
\cdot
\end{align*}
To show the effect of the AR update, we make simplifying assumptions to derive the update for the conditional mean function, $\mu_i(x) = \int y \, p_i(y \mid x) dy$. Let us assume that our predictive densities are normally distributed, that is $P_{i-1}(y \mid x) = \mathcal{N}(y \mid \mu_{i-1}(x), \sigma_y^2)$. This is an accurate approximation if the truth is normal and we have observed sufficient observations. Without loss of generalizability, we assume that $\sigma_y^2 = 1$. This then gives the form $P_{i-1}(y \mid x) = \Phi(y-\mu_{i-1}(x))$, which will help us in the calculation of the bivariate Gaussian copula. If we multiply by $y$ and integrate on both sides of \eqref{eq:app_condit}, we get
\begin{align*}
\mu_i(x) = [1-\alpha_i(x,x_i)]\mu_{i-1}(x)  + \alpha_i(x,x_i) \int \, c\left(P_{i-1}(y \mid x), P_{i-1}( y_i \mid x_i); \rho \right)\, y \, p_{i-1}(y \mid x) \, dy.
\end{align*}
Plugging in $P_{i-1}(y \mid x) = \Phi\{y-\mu_{i-1}(x)\}$ (and similarly for the density) to the above gives
\begin{align*}
 \int c\left(P_{i-1}(y \mid x), P_{i-1}( y_i \mid x_i);\rho \right)\, y \, dy &=\int \frac{\mathcal{N}(y, y_i \mid  [\mu_{i-1}(x),\mu_{i-1}(x_i)], 1,\ \rho)}{\mathcal{N}(y_i   \mid \mu_{i-1}(x_i),1)} \, \,y \, dy\cdot
\end{align*}
The above  is simply the expectation of a conditional normal distribution, giving us
\begin{align*}
\int c\left(P_{i-1}(y \mid x), P_{i-1}( y_i \mid x_i) ; \rho \right)\, y \, dy = \mu_{i-1}(x) + \rho(y_i - \mu_{i-1}(x_i)).
\end{align*}
Putting it all together, we thus have
\begin{align*}
\mu_i(x) =\mu_{i-1}(x) + \alpha_i(x,x_i) \rho(y_i - \mu_{i-1}(x_i)).
\end{align*}
In the autoregressive case, we have
\begin{align*}
\mu_i(x) =\mu_{i-1}(x) +  \alpha_i(x,x_i) \rho(x, x_i) (y_i - \mu_{i-1}(x_i)),
\end{align*}
where we use the notations $\rho_i(x)=\rho( x,x_i)$ interchangeably to highlight the dependence of $\rho$ on the distance between $x$ and $x_i$.
Further assuming $P_{i-1}(x) = \mathcal{N}(x \mid 0,1)$ returns a tractable form for $\alpha_i(x,x')$, giving us Figure \ref{fig:itupd} in the main paper.

\subsection{Derivation of Copula Update for Supervised Learning} \label{app:deriv_suplearn}
We now derive the predictive density update for supervised learning tasks, closely following the derivations of \citet{fong2021martingale} for the conditional methods in Supplements E.2 and E.3.
We assume fixed design points ${x}_{1:n}\in\mathbb{R}^{n\times d}$ and random response $y_{1:n}\in\mathbb{R}^{n}$.

\subsubsection{Conditional Regression with Dependent Stick-Breaking}\label{app:deriv_copula_regression}
We follow Appendix E.2.2 in \cite{fong2021martingale}, and derive the regression copula update inspired by the dependent DP. Consider the general covariate-dependent stick-breaking mixture model 
\begin{equation}\label{SM_eq:DDP_mixture_location}
\begin{aligned}
f_{G_{x}}({y}) = \int \mathcal{N}(y \mid \theta,1) \, dG_{x}(\theta), \quad
G_{x} =\sum_{l =1}^\infty w_l(x)\,\delta_{\theta^*_l(x)}.
\end{aligned}
\end{equation}
For the weights, we elicit the stick-breaking prior $w_l(x) = v_l(x) \prod_{j< l} \{1-v_j(x)\}$ where $v_l(x)$ is a stochastic process on $\mathcal{X}$ taking values in $[0,1]$, and is independent across $l$. For the atoms, which are now dependent on $x$, we assume they are independently drawn from a Gaussian process,
\begin{align*}
\theta_l^*(\cdot) \iid  \text{GP}(0, \tau^{-1} k),
\end{align*}
where $k$ is the covariance function. Once again, we want to compute
\begin{align*}
\frac{E\left[f_{G_{x}}({y}) \, f_{G_{x_1}}({y}_1) \right]}{p_0(y \mid x) \, p_0(y_1 \mid x_1)}\cdot
\end{align*}
Following the stick-breaking argument as in Section \ref{Appendix:multivariate}, we can write the numerator as
\begin{align*}
\left\{1-\beta_1(x,x_1)\right\}E\left[ K\{{y} \mid \theta^*(x)\} \right] E\left[ K\{{y}_1 \mid \theta^*(x_1)\} \right] + \beta_1(x,x_1) E\left[ K\{{y} \mid \theta^*(x)\} \, K\{{y}_1 \mid \theta^*(x_1)\} \right] 
\end{align*}
where
\begin{align*}K\{y \mid \theta^*(x)\} = \mathcal{N}\{y \mid \theta^*(x),1\}, \quad \theta^*(\cdot) \sim \text{GP}(0, \tau^{-1} k),\end{align*} 
and
\begin{align*}
\beta_1(x,x_1) = \sum_{k=1}^\infty E\left[ w_k(x)w_k(x_1) \right].
\end{align*}
As before, we have
\begin{align*}
\frac{E\left[f_{G_{x}}({y}) \, f_{G_{x_1}}({y}_1) \right]}{p_0(y \mid x) \, p_0(y_1 \mid x_1)} = c\left\{ P_0(y \mid x), P_0(y_1 \mid x_1); \rho_1(x)\right\}
\end{align*}
where $\rho_1(x) = \rho_0 k(x,x_1)$ and $\rho_0 = 1/(1+\tau)$. We thus have the copula density as a mixture of the independent and Gaussian copula density. This then implies the copula update step of the form
\begin{equation*}
\begin{aligned}
p_{i}(y \mid x) = \left[1-\beta_i(x,x_i)+ \beta_i(x,x_i) \, c\left\{P_{i-1}(y \mid x),P_{i-1}(y_i \mid x_i); \rho_i(x)\right\}\right]\,  p_{i-1}(y\mid x),
\end{aligned}
\end{equation*}
where we write $\rho_i(x)=\rho^{d+1}_i(x)$.
As in \cite{fong2021martingale}, we turn to the multivariate update for inspiration where we do not update $P_n(x)$ and instead keep it fixed at $P_0(x) = \Phi(x)$ (for each dimension). This gives us 
\begin{align} \label{eq:condit_alpha}
\beta_{i}({x},{x}_{i}) = \frac{\alpha_{i}\prod_{j=1}^d  c\left\{\Phi\left(x^j\right),\Phi\left(x_{i}^j\right); \rho^{j}_i(x^{1:j-1})\right\}}{1- \alpha_{i} + \alpha_{i}\prod_{j=1}^d   c \left\{\Phi\left(x^j\right),\Phi\left(x_{i}^j\right); \rho^{j}_i(x^{1:j-1})\right\}}\cdot
\end{align}

\subsubsection{Classification with Beta-Bernoulli Copula Update}\label{app:deriv_copula_classification} 
In the classification setting (Appendix E.3.1 in \cite{fong2021martingale}), \citet{fong2021martingale} assume a beta-Bernoulli mixture for $y_i \in \{0,1\}$. As the derivation is written w.r.t  $\rho$, we simply replace $\rho$ with our definition of $\rho^{j}_i(x^{1:j-1})$, giving the update
\begin{align*}
p_{i}(y \mid {x}) = \left(1-\beta_{i}({x},{x}_{i})+ \beta_{i}({x},{x}_{i})\, b\left\{q_{i-1},r_{i-1}; \rho_i(x)\right\}\right) p_{i-1}(y\mid {x})
\end{align*}
where $q_{i-1} = p_{i-1}(y \mid {x}),r_i = p_{i-1}(y_{i}\mid {x}_{i})$, $\rho_i(x)$ as in Equation \ref{eq:rho}, $\beta_{i}({x},{x}_{i})$ similarly as in \eqref{eq:condit_alpha}, and finally the copula-like function $b$ given by
\begin{align*}
b\{q_{i-1},r_{i-1}; \rho_i(x)\} &= \begin{dcases} 
    1-\rho_i(x) +\rho_i(x)\,\frac{q_{i-1}\wedge r_{i-1}}{q_{i-1}\, r_{i-1}} &\quad \text{if } y = y_{i} \vspace{2mm}\\  
    1-\rho_i(x) + \rho_i(x)\,\frac{q_{i-1} - \{q_{i-1}\wedge (1-r_{i-1})\}}{q_{i-1}\, r_{i-1}}&\quad \text{if } y \neq y_{i} \cdot
\end{dcases}
\end{align*}

\section{Methodology} \label{app:algs}
In this section, we provide more details on the methodology referred to in the main part of the paper.
\subsection{Generative Modelling} \label{app:sampling}
First, we consider three approaches to generative modelling
\begin{enumerate}
    \item Inverse sampling
    \item Importance sampling
    \item \gls{smc}
\end{enumerate}

\subsubsection{Inverse Sampling}

\paragraph{Univariate setting}
As noted by \citet{fong2021martingale}, we can sample from $x^*\sim P_n(x)$ by inverse sampling, that is 
\begin{align*}
    u    \sim \mathcal{U}[0,1],\;\; x^*  \sim P_n^{-1}(u).
\end{align*}
As we cannot evaluate $P_n^{-1}(u)$ directly, we instead solve an optimisation problem
\begin{align*}
    x^* = \argmin_x |P_n(x) - u| %
\end{align*}

\paragraph{Multivariate setting} 
The univariate procedure can be repeated iteratively in the multivariate setting given the conditional distribution
\begin{align*}
    u^1 &\sim \mathcal{U}[0,1], \;x^1 = P_n^{-1}(u^1)\\
    u^2 &\sim \mathcal{U}[0,1], \;x^2 = P_n^{-1}(u^2 \mid x^1)\\
    \ldots\\
    u^d &\sim \mathcal{U}[0,1], \;x^d= P_n^{-1}(u^d \mid x^{1:d-1})\\    
\end{align*}

\subsubsection{Importance Sampling}
In practice, inverse sampling is unstable and is highly dependent on the performance of the optimization. An alternative approach to data generation is importance sampling. This includes two steps
\begin{enumerate}
    \item Sampling a set of particles $z_1,\ldots,z_{B}$ from the initial predictive $p_0$.
    \item Re-sampling $z_1,\ldots,z_{B}$ with replacement based on the weights $w_1=p_n(z_1)/p_0(z_1),\ldots,w_B=p_n(z_B)/p_0(z_B)$.
\end{enumerate}

\subsubsection{Sequential Monte Carlo}
Importance sampling will perform poorly if $p_n$ and $p_0$ are far apart. Instead, we propose a \gls{smc} procedure. A similar \gls{smc} sampling scheme has been proposed for univariate imputation of censored survival data by \citet{fong2022predictive}. Here, the goal is parameter inference, and thus only requires \textit{implicit} sample observations by drawing the marginal \gls{cdf} $u_n^j$ from a uniform distribution. In our case, we generate new \textit{explicit} data directly by sampling from the data space. Please see Algorithm \ref{alg:sample} for a complete overview. As this sampling approach is similar to evaluating the density at test data points (Algorithm \ref{alg:eval_density}), we highlighted the differences in blue. In short,
\begin{enumerate}
    \item We sample a set of particles $z_1,\ldots,z_{B}$ from the initial predictive $p_0$, and set the particle weights to $w^{[0]}_k=1$ for all $k=1,\ldots, B$
    \item We update the predictive $p_{i-1}\rightarrow p_{i}$, and the particle weights $w^{[i]}_k=w^{[i-1]}_k\cdot p_{i}{\left(z^{[i-1]}_{k}\right)}/p_{i-1}{\left(z^{[i-1]}_{k}\right)}$ for each training observation
    \item If the \gls{ess} is smaller than half of the number of particles, we resample $z_1,\ldots,z_{B}$ and $w^{[i]}_1,\ldots,w^{[B]}_1$ based on their weights.
\end{enumerate}
Note that particle diversity can be improved by introducing move steps, for example using Markov kernels  \cite{chopin2002sequential, gunawan2020subsampling}.

\begin{figure}
    \centering
    \centering
    \subfloat[Inverse Sampling]{\includegraphics[height = \textwidth/5]{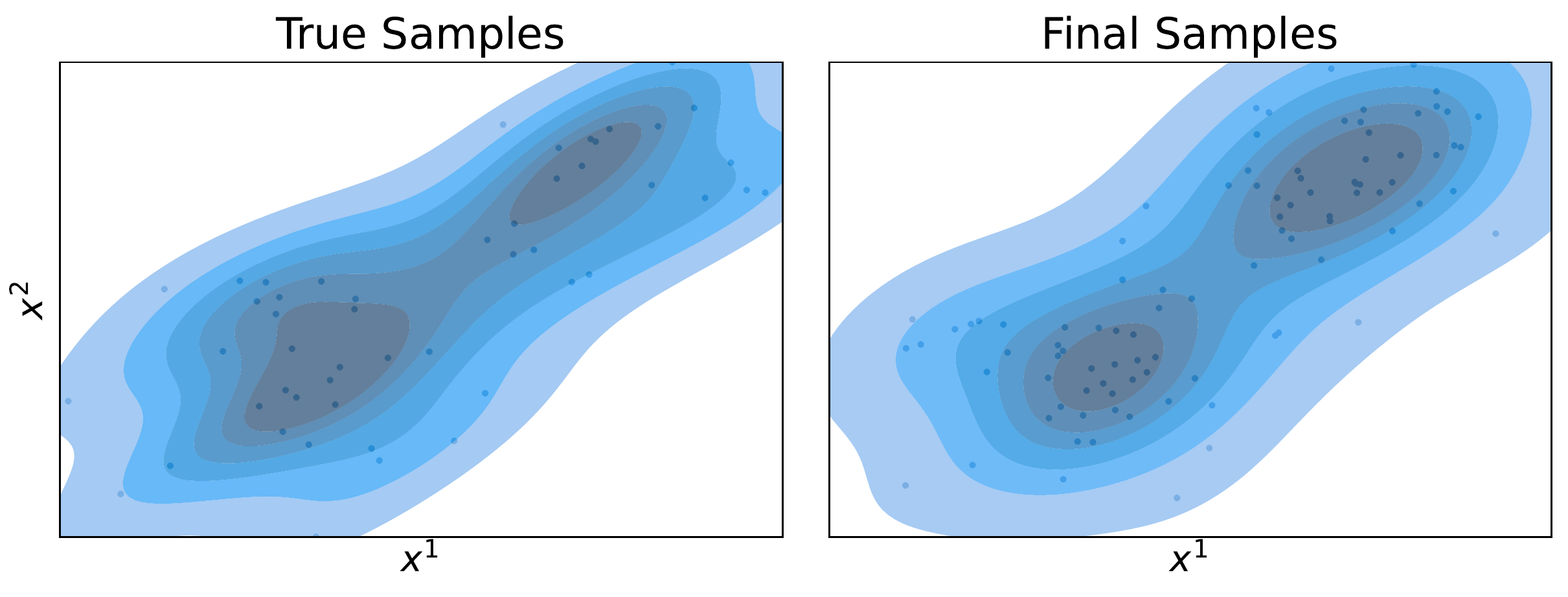}}\\
    \subfloat[Importance Sampling]{\includegraphics[height = \textwidth/5]{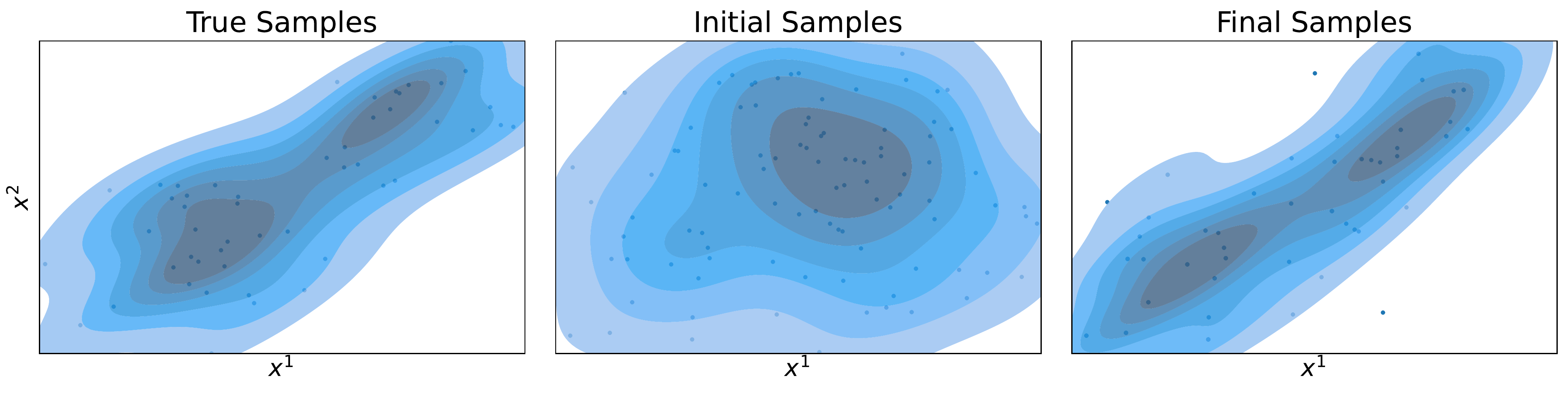}}\\
    \subfloat[Sequential Monte Carlo]{\includegraphics[height = \textwidth/5]{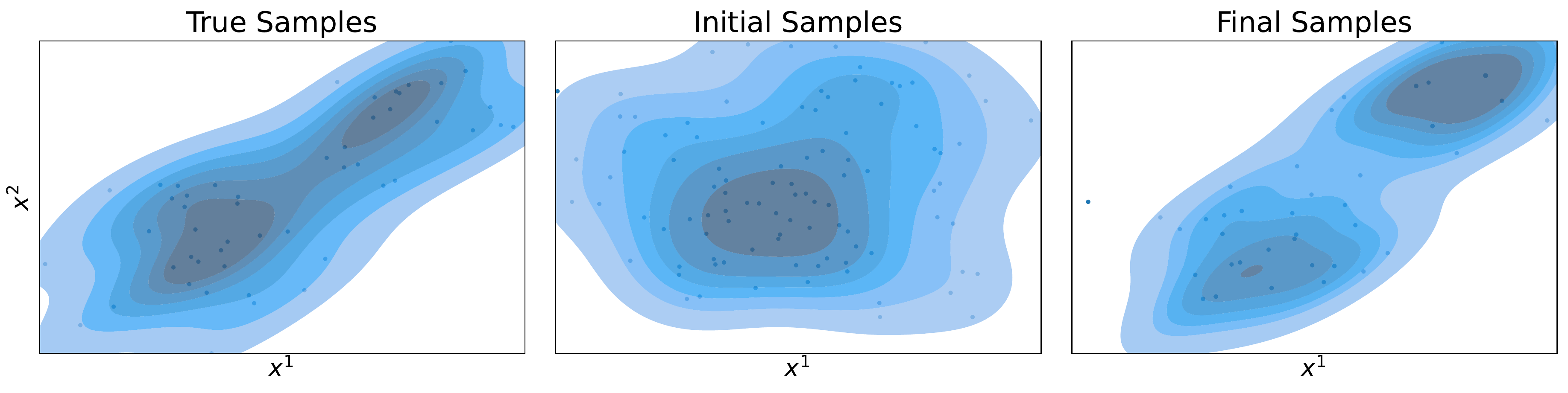}}
    \caption{100 samples generated from AR$_d$-BP trained on 50 samples from a GMM with 4 components. All three sampling approaches manage to preserve the multi-modal data distribution.}
    \label{fig:gmm-samples}
\end{figure}

In Figure \ref{fig:gmm-samples}, we see that inverse sampling struggles on a simple GMM example. On the other hand, importance sampling and SMC provide reasonable samples.
Similar sampling schemes have been proposed for Restricted Boltzmann Machines \citep{larochelle2011neural, salakhutdinov2008quantitative} where samples can only be drawn from the model approximately by Gibbs sampling.

\subsection{Supervised Learning} \label{app:supervised}
We briefly recap how joint density estimation can be extended to conditional supervised learning (regression and classification), as outlined by \citet{fong2021martingale}. Please see Supplement \ref{app:deriv_suplearn} for the derivation.
Given fixed design points ${x}_{1:n}$ and random response $y_{1:n}$, the problem at hand is to infer a {family} of conditional densities $\{f_{x}(y): {{x} \in \mathbb{R}^d}\}$.

\subsubsection{Regression} \label{sec:conditreg}
For the regression case, \citet{fong2021martingale} posit a Bayesian model with the nonparametric likelihood being a covariate-dependent stick-breaking \gls{dpmm}:
\begin{align}\label{eq:DDP_mixture}
f_{G_{x}}({y}) = \int \mathcal{N}(y \mid \theta,1) \, dG_{{x}}({\theta}), \quad G_{x} =\sum_{k=1}^\infty w_k({x})\,\delta_{\theta_k^*},
\end{align}
where $w_k(\mathbf{x})$ follows an ${x}$-dependent stick-breaking process. Our contribution is to assume an autoregressive factorisation of the kernel and independent GP priors on $\theta_k^*$. See Supplement \ref{app:deriv_copula_regression} for the derivation of the predictive density update that is now given by
\begin{align} \label{eq:update_p_reg}
p_{i}(y \mid x) = \left[1-\beta_i(x,x_i)+ \beta_i(x,x_i) \, c\left\{P_{i-1}(y \mid x),P_{i-1}(y_i \mid x_i); \rho_i(x)\right\}\right]\,  p_{i-1}(y\mid x),
\end{align}
where $\rho_i(x)=\rho^{d+1}_i(x)$ and $\beta$ as in \eqref{eq:condit_alpha}.

\subsubsection{Classification}
For $y_i \in \{0,1\}$, \citet{fong2021martingale} assume a beta-Bernoulli mixture. As explained in Supplement {\ref{app:deriv_copula_classification}} and \cite{fong2021martingale}, this gives the same update as in the regression setting with the difference that the copula $c$ in \eqref{eq:update_p_reg} is replaced with 
\begin{align*}
b\{q_{i-1},r_{i-1}; \rho_i(x)\} &= \begin{dcases} 
    1-\rho_i(x) +\rho_i(x)\,\frac{q_{i-1}\wedge r_{i-1}}{q_{i-1}\, r_{i-1}} &\quad \text{if } y = y_{i} \vspace{2mm}\\  
    1-\rho_i(x) + \rho_i(x)\,\frac{q_{i-1} - \{q_{i-1}\wedge (1-r_{i-1})\}}{q_{i-1}\, r_{i-1}}&\quad \text{if } y \neq y_{i},
\end{dcases}
\end{align*}
where $\rho_i(x)=\rho^{d+1}_i(x), q_{i-1} = p_{i-1}(y \mid \mathbf{x}),r_{i-1} = p_{i-1}(y_{i}\mid \mathbf{x}_{i})$ and $\rho_y \in (0,1)$. 

\subsection{Implementation Details} \label{app:imp}
Please see Algorithm \ref{alg:full_alg} for the full estimation procedure, Algorithm \ref{alg:optimal_rho} for the optimisation of the bandwidth parameters, Algorithm \ref{alg:fit_density} for the fitting procedure of the predictive density updates, and eventually Algorithm \ref{alg:eval_density} for the steps during test-time inference. All algorithms are written for one specific permutation of the dimensions, and are repeated for different permutations.

Note that at both training time and test time, we need to consider the updates on the scale of the \gls{cdf}s, that is for the terms such as $u_i^j(x^j)$, which appear in the update step \eqref{eq:mv_DP_AR_marginal}.
Given 
\begin{align*}
u_{i}^{j}(x^j)=P_i(x^j|x^{1:j-1}) = \int_{-\infty}^{x^j} {p_i(x^{1:j-1}, x^{'j})}/{p_i(x^{1:j-1})} dx^{'j},
\end{align*}
and \eqref{eq:mv_DP_AR_marginal}, the \gls{cdf}s $u_{i}^{j}(x^{j})$ take on the tractable update %
\begin{align} \label{eq:mv_autoreg_DP_copdistr}  %
\displaystyle{
u_{i}^{j} =\left\{(1-\alpha_{i})u_{i-1}^j + \alpha_{i} H\left(u_{i-1}^j,v_{i-1}^j; \rho_i^j \right) \prod_{r=1}^{k-1} c\left(u_{i-1}^{r},v_{i-1}^{r}; \rho_i^r\right)\right\} \frac{p_{i-1}\left(x^{1:k-1}\right)}{p_{i}\left(x^{1:k-1}\right)}},
\end{align}
and set $v_{i-1}^{j} = u_{i-1}^{j}(x_{i})$ which holds by definition, where we dropped the argument $x$ for simplicity from $\rho_i^j$ and $u_{i}^{j}$, and $H(u,v; \rho)$ denotes the conditional Gaussian copula distribution with correlation $\rho$, that is
\begin{align*}
    H(u,v; \rho) = \int_0^u c(u',v; \rho) du' = \Phi\left\{\frac{\Phi^{-1}(u) -\rho \Phi^{-1}(v)}{\sqrt{1-\rho^2}} \right\}\cdot
\end{align*}
The Gaussian copula density $c(u,v; \rho)$ is given by
\begin{align*}
c(u,v; \rho) = \frac{\mathcal{N}_2\left\{ \Phi^{-1}(u),\Phi^{-1}(v)\mid 0,1,\rho\right\}	}{\mathcal{N}\{\Phi^{-1}(u) \mid 0,1\} \mathcal{N}\{\Phi^{-1}(v) \mid 0,1\}} ,
\end{align*}
where $\Phi$ is the normal \gls{cdf}, and $\mathcal{N}_2$ is the standard bivariate density with correlation $\rho \in (0,1)$. 

\paragraph{Ordering} \label{app:ordering}
Note that the predictive density update depends on the ordering of both the training data and the dimensions.
This permutation dependence is not an additional assumption on the data generative process, and the only implication is that the subset of ordered marginal distributions continue to satisfy \eqref{eq:mv_DP_marginal} (main paper).
In the absence of a natural ordering of the training samples or the dimensions, we take multiple random permutations, observing in practice that the resulting averaged density estimate performs better.
More precisely, for a given permutation of the dimensions, we first tune the bandwidth parameters, and then calculate density estimates based on multiple random permutations of the training data. 
We then average over each of the resulting estimates to obtain a single density estimate for each dimension permutation, and subsequently take the average across these estimates to obtain the final density estimate. 
Importantly, our method is parallelizable over permutations and thus able to exploit modern multi-core computing architectures.

\begin{figure}[ht]
\small
\centering
\begin{minipage}{\linewidth}
\begin{algorithm}[H]
\caption{Full density estimation pipeline}
{\textbf{Input:}{\\
\hspace*{\algorithmicindent} \train: training observations;\\
\hspace*{\algorithmicindent} \test: test observations;\\
\hspace*{\algorithmicindent} $M$: number of permutations over samples and features to average over;\\
\hspace*{\algorithmicindent} $n_\rho$: number of train observations used for the optimisation of bandwidth parameters;
  }}
\textbf{Output:}\\ 
\hspace*{\algorithmicindent} {$p_{\nr}(x_{\nr+1}), \ldots p_{\nr}(x_{\nr+\ns})$}: density of test points \\
\begin{algorithmic}[1]
\Procedure{full_density_estimation}{}
\State Compute optimal bandwidth parameters  \graycomment{$\mathcal{O}(Mn_{\rho}^2\numfeats\cdot \#\text{gradient steps})$}  \vspace{2mm}
\State Compute $v_i^{j, (m)}$ for $i\in\{1,\ldots,\nr\}, j\in\{1,\ldots,\numfeats\}, m\in\{1,\ldots, M\}$ \graycomment{$\mathcal{O}(M\nr^2\numfeats)$}  \vspace{2mm}
\State Evaluate density at test observations \test \graycomment{$\mathcal{O}(M\nr\ns\numfeats)$} 
\EndProcedure
\end{algorithmic}\label{alg:full_alg}
\end{algorithm}
\end{minipage}
\end{figure}

\begin{algorithm}[H]
\caption{Estimate optimal bandwidth parameters}
{\textbf{Input:}{\\
\hspace*{\algorithmicindent} \train: training observations;\\
\hspace*{\algorithmicindent} $M$: number of permutations over samples and features to average over;\\
\hspace*{\algorithmicindent} $n_\rho$: number of train observations used for the optimisation of bandwidth parameters;\\ \vspace{2mm}
\hspace*{\algorithmicindent} \texttt{maxiter}: number of iterations;\\
\hspace*{\algorithmicindent} $\mathcal{R}^{(0)}$: initialisation of bandwidth parameters: \\
\hspace*{\algorithmicindent} -$\mathcal{R}^{(0)}=\{\rho^{(0)}_0, l^{(0)}_1, \ldots, l^{(0)}_{d-1}\}$ (by default, $\rho^{(0)}_0\gets 0.9, l^{(0)}_1\gets1, \ldots, l^{(0)}_{d-1}\gets1$) for AR-BP, \\ 
\hspace*{\algorithmicindent} -$\mathcal{R}^{(0)}=\{\rho^{(0)}_0, w\}$ (by default, $\rho^{(0)}_0\gets 0.9$, and $w$ initialised as implemented in \texttt{Haiku} by default) for ARnet-BP
}} \\ 
\textbf{Output:}\\ 
\hspace*{\algorithmicindent} $\mathcal{R}^{(\text{\texttt{maxiter}})}$: optimal bandwidth parameters \\ 
\begin{algorithmic}[1]

\Procedure{optimal_bandwidth_and_lengthscales}{}
\State Subsample $\{x'_1, \ldots, x'_{n_\rho}\}$ from \train %
\For{$s \gets 1$ \textnormal{\textbf{to}} \texttt{maxiter}}
\State _, $\{p_{i-1}^{(m)}\left(x'_i\right)\}_{i, m}\gets$ \textsc{fit_conditional_predictive_cdf}(
\Statex \hskip60pt{$\mathcal{R}^{(s-1)}, \{x'_1, \ldots, x'_{n_\rho}\}, M$, fit_density=True)} \vspace{4mm} %
\State Compute $L(x'_1, \ldots, x'_{n_\rho}) = -\sum_{m=1}^M\sum_{i=1}^{n_\rho} \log p^{(m)}_{i-1}(x'_{i})$ \vspace{4mm}
\State \indentbox{$\mathcal{R}^{(s)}\gets$ \textsc{adam_step}($\mathcal{R}^{(s-1)}$, $L$)} %
\EndFor\\
\hspace*{4mm} \Return $\mathcal{R}^{(s)}$
\EndProcedure
\end{algorithmic}\label{alg:optimal_rho}
\end{algorithm}

\begin{algorithm}[H]
\caption{Single copula update}
{\textbf{Input:}{\\
\hspace*{\algorithmicindent} $z$: observation to update the log density;\\
\hspace*{\algorithmicindent} $x_i$: observation to update with;\\
\hspace*{\algorithmicindent} $i$: sample index;\\
\hspace*{\algorithmicindent} $j$: feature index;\\
\hspace*{\algorithmicindent} $u_{i-1}^j(z)$: predictive \gls{cdf} for $z$;\\
\hspace*{\algorithmicindent} $v_{i-1}^{j, (m)}$: prequential \gls{cdf};\\
\hspace*{\algorithmicindent} $\rho^j_{i}(z^{1:j-1})$=None: bandwidth;\\
\hspace*{\algorithmicindent} $\mathcal{R}$=None: bandwidth parameters;\\
}}
\textbf{Output:}\\ 
\hspace*{\algorithmicindent} \indentbox{$u_i(z)$} \\ 

\begin{algorithmic}[1]
\Procedure{CDF_Update}{}
\State \parbox[t]{313pt}{Compute
\begin{equation*}
    \rho^j_{i}(z^{1:j-1}) \gets \rho_0 k_{\mathcal{R}}\left(z^{1:j-1}, x^{1:j-1}_i\right)
\end{equation*}
where $k_{\mathcal{R}}$ denotes the user-defined kernel if $\rho=$None \strut} \vspace{4mm}
\State \parbox[t]{313pt}{Compute the bivariate Gaussian copula density
\begin{multline*}
    c\{u_{i-1}^j(z), v_{i-1}^{j, (m)}; \rho_j\} \gets \frac{\mathcal{N}_2\left\{ \Phi^{-1}(u_{i-1}^j(z)),\Phi^{-1}(v_{i-1}^{j, (m)})\mid 0,1,\rho^j_i(z^{1:j-1})\right\}	}{\mathcal{N}\{\Phi^{-1}(u_{i-1}^j(z)) \mid 0,1\}\, \mathcal{N}\{\Phi^{-1}(v_{i-1}^{j, (m)}) \mid 0,1\}} %
\end{multline*} \strut}
\State \parbox[t]{320pt}{Compute the conditional Gaussian copula {CDF}
\begin{multline*}
H\left\{u_{i-1}^j(z), v_{i-1}^{j, (m)}; {\rho^j_i(z^{1:j-1})}\right\}\gets \Phi\left\{ \frac{\Phi^{-1}(u_{i-1}^j(z)) - \rho^j_i(z^{1:j-1}) \Phi^{-1}(v_{i-1}^{j, (m)})}{\sqrt{1-\rho^j_i(z^{1:j-1})^2}}\right\} 
\end{multline*} \strut}
\State Compute $\alpha_i=(2-\frac{1}{i})\frac{1}{i+1}$
\State \parbox[t]{313pt}{Compute $u_{i}^j(z) = P_{i}^j(z|z^{1:j-1})$ by
\begin{multline*} 
      \displaystyle u_{i}^{j}(z) \gets  \Bigg\{(1-\alpha_{i})u_{i-1}^j(z) + 
      \displaystyle \alpha_{i} H\Bigg(u_{i-1}^j(z), v_{i-1}^{j, (m)}; {\rho^j_{i}(z)} \Bigg) \prod_{l=1}^{j-1} c\Bigg(u_{i-1}^l(z), v_{i-1}^{l, (m)}; {\rho^j_{i}(z)}\Bigg)\Bigg\}\\   \;\;\cdot1\Bigg/\Bigg\{1-\alpha_{i} + \alpha_{i} \prod_{j=1}^d {c\Bigg(u_{i-1}^{j}(z), v_{i-1}^{j, (m)}; {\rho^j_{i}(z)}\Bigg)}\Bigg\}
\end{multline*}
\strut}\\
\hspace*{4mm} \Return {$u_i(z)$} 
\EndProcedure
\end{algorithmic}\label{alg:copula_update}
\end{algorithm}

{

\begin{algorithm}[H]
\caption{Estimate prequential CDFs at train observations}
{\textbf{Input:}{\\
\hspace*{\algorithmicindent} $\mathcal{R}$: bandwidth parameters\\
\hspace*{\algorithmicindent} \train: training observations;\\
\hspace*{\algorithmicindent} $M$: number of permutations over features to average over;\\
\hspace*{\algorithmicindent} compute_density (by default, False);\\
}}
\textbf{Output:}\\ 
\hspace*{\algorithmicindent} $\{v_{i-1}^{j, (m)}\}_{i, j, m}, \{p_{i-1}^{(m)}\left(x_i\right)\}_{i, m}$ {if} compute_density, {else} $\{v_{i-1}^{j, (m)}\}_{i, j, m}$  \\ 
\begin{algorithmic}[1]
\Procedure{fit_conditional_predictive_cdf}{}
\For{$m \gets 1$ \textnormal{\textbf{to}} $M$}
    \State \indentbox{Sample permutation $\pi_1\in\Pi(\nr), \pi_2\in\Pi(d)$ } \vspace{4mm}
    \State \parbox[t]{313pt}{%
    Change the ordering of the training observations $\{x^{(m)}_1,\ldots,x^{(m)}_{\nr}\} \gets \{\pi_1(x_1),\ldots, \pi_1(x_{\nr})\}$ and the features $x \gets [\pi_2(x^1),\ldots,\pi_2(x^d)]$ \graycomment{For simplicity we will drop the superscript in the following}} \vspace{4mm}
    \For{$j \gets 1$ \textnormal{\textbf{to}} $\numfeats$}
        \For{$k \gets 1$ \textnormal{\textbf{to}} $\nr$}
            \State \parbox[t]{313pt}{Initialise ${u_{0}^j(x_{k})} \gets \Phi(x_{k}^j)$ \graycomment{$u$ also depends on the permutation $m$, but since we do not reuse $u$ after $m$ is updated, we drop the index for simplicity}\strut}  
        \EndFor
    \EndFor \vspace{3mm}
    \For{$i \gets 1$ \textnormal{\textbf{to}} ${\nr}$}
        \State Set $v_{i-1}^{j, (m)} \gets u_{i-1}^{j}(x_{i})$ \textbf{for} $j \gets 1 \textnormal{ \textbf{to} } \numfeats$
        \For{$k \gets 1$ \textnormal{\textbf{to}} $i$}
            \For{$j \gets 1$ \textnormal{\textbf{to}} $\numfeats$}
                \State $u_{i}^j(x_k)$=\textsc{cdf_update}($x_k$, $x_i$, $i$, $j$, $u_{i-1}^j(x_k)$, $v_{i-1}^{j, (m)}$, $\mathcal{R}$)

            \EndFor
            \If {{compute_density}}
                \State \parbox[t]{313pt}{
                \begin{equation*}
                    p_{i}^{(m)}{\left(x_k\right)} \gets \left\{1-\alpha_{i} + \alpha_{i} \prod_{j=1}^d {c\left(u_{i-1}^{j}(x_k), v_{i-1}^{j, (m)}; \rho^j_{i}(x_k)\right)}\right\} p_{i-1}^{(m)}\left(x_k\right)
                \end{equation*} \strut}
            \EndIf            
        \EndFor
    \EndFor
\EndFor\\
\hspace*{4mm} \Return {$\{v_{i-1}^{j, (m)}\}_{i, j, m}, \{p_{i-1}^{(m)}\left(x_i\right)\}_{i, m}$} \textbf{if} compute_density \textbf{else} $\{v_{i-1}^{j, (m)}\}_{i, j, m}$ 
\EndProcedure
\end{algorithmic}\label{alg:fit_density}
\end{algorithm}
}

{

\begin{algorithm}[H]
\caption{Evaluate density at test observations}
{\textbf{Input:}{\\
\hspace*{\algorithmicindent} $\mathcal{R}$: bandwidth parameters\\
\hspace*{\algorithmicindent} \test: test observations;\\
\hspace*{\algorithmicindent} $\{\{x^{(1)}_1,...,x^{(1)}_{\nr}\},\ldots,\{x^{(M)}_1,...,x^{(M)}_{\nr}\}\}$: sets of permuted train observations;\\
\hspace*{\algorithmicindent} $\{v^{j, (m)}_i\}_{i, j, m}$: {prequential conditional CDFs at train observations};\\
\hspace*{\algorithmicindent} $M$: number of observations over features to average over;\\
}}
\textbf{Output:}\\ 
\hspace*{\algorithmicindent} \indentbox{$\{p_\nr(x_{\nr + 1}), \ldots, p_\nr(x_{\nr + \ns}
)\}$%
}
\\ 
\begin{algorithmic}
\Procedure{eval_density}{}
\For{$m \gets 1$ \textnormal{\textbf{to}} $M$}
    \For{$j \gets 1$ \textnormal{\textbf{to}} $\numfeats$}
        \For{$k \gets 1$ \textnormal{\textbf{to}} {$\ns$}}
            \State \parbox[t]{313pt}{Initialise ${u_{0}^j(x_{\nr+k})} \gets \Phi(x_{\nr+k}^j)$}
        \EndFor
    \EndFor \vspace{3mm}
    \For{$i \gets 1$ \textnormal{\textbf{to}} ${\nr}$}
        \For{$k \gets 1$ \textnormal{\textbf{to}} ${\ns}$}
            \For{$j \gets 1$ \textnormal{\textbf{to}} $\numfeats$}
                \State $u_{i}^j(x_k)$=\textsc{cdf_update}($x_{\nr + k}$, $x_i$, $i$, $j$, $u_{i-1}^j(x_{\nr + k})$, $v_{i-1}^{j, (m)}$, $\mathcal{R}$)

            \EndFor
            \State \parbox[t]{313pt}{ Compute density
            \begin{multline*}
            p_{i}^{(m)}{\left(x_{\nr + k}\right)} \gets\\ \left\{1-\alpha_{i} + \alpha_{i} \prod_{j=1}^d {c\left(u_{i-1}^{j}(x_{\nr+ k}), v_{i-1}^{j, (m)}; \rho^j_{i}(x_{\nr + k})\right)}\right\} p_{i-1}^{(m)}\left(x_{\nr + k}\right)
            \end{multline*} \strut}
        \EndFor
    \EndFor
\EndFor\\
\State\textcolor{gray}{$\triangleright$ Average density over permutations}
\For{$i \gets \nr+1$ \textnormal{\textbf{to}} $\nr+\ns$}
\State $p_{\nr}(x_{i}) \gets \frac{1}{M}\sum_{m=1}^{M} p^{(m)}_{\nr}(x_{i})$ 
\EndFor\\
\hspace*{4mm} \Return {$\{p_{\nr}(x_{\nr + 1}), \ldots, p_{\nr}(x_{\nr + \ns})\}$} 
\EndProcedure
\end{algorithmic}\label{alg:eval_density}
\end{algorithm}
}

\begin{algorithm}[H]
\caption{Sample new observations}
{\textbf{Input:}{\\
\hspace*{\algorithmicindent} $\mathcal{R}$: bandwidth parameters\\
\hspace*{\algorithmicindent} $\{z^{[0]}_1, \ldots, z^{[0]}_{\np}\}$: initial samples from proposal distribution;\\
\hspace*{\algorithmicindent} $\{q(z^{[0]}_1), \ldots, q(z^{[0]}_{\np})\}$: proposal density evaluated at initial samples;\\
\hspace*{\algorithmicindent} $\{\{x^{(1)}_1,...,x^{(1)}_{\nr}\},\ldots,\{x^{(M)}_1,...,x^{(M)}_{\nr}\}\}$: sets of permuted train observations;\\
\hspace*{\algorithmicindent} $\{v^{j, (m)}_i\}_{i, j, m}$: prequential conditional CDFs at train observations;\\
}}\\
\textbf{Output:}\\ 
\hspace*{\algorithmicindent} \indentbox{$\{z^{[{\nr}]}_1, \ldots, z^{[{\nr}]}_{\np}\}$ and $\{p_\nr(z^{[{\nr}]}_1), \ldots, p_\nr(z^{[{\nr}]}_{\np})\}$}\\ 
\begin{algorithmic}[1]
\Procedure{sample}{}
\For{$m \gets 1$ \textnormal{\textbf{to}} $M$}
    \For{$k \gets 1$ \textnormal{\textbf{to}} {$\np$}}
        \For{$j \gets 1$ \textnormal{\textbf{to}} $\numfeats$}
            \State {Initialise ${u_{0}^j(z^{[0]}_k)} \gets \Phi(z^{[0]}_k)$}
        \EndFor
        \State Initialise ${w^{[0]}_{k}} \gets p_{0}(z^{[0]}_k)/q(z^{[0]}_{k})$ %
    \EndFor 
\EndFor \vspace{3mm}
\For{$i \gets 1$ \textnormal{\textbf{to}} ${\nr}$}
    \For{$k \gets 1$ \textnormal{\textbf{to}} ${\np}$}
        \For{$m \gets 1$ \textnormal{\textbf{to}} $M$}
            \For{$j \gets 1$ \textnormal{\textbf{to}} $\numfeats$}
                \State $u_{i}^j(x_k)$=\textsc{cdf_update}($z^{[i-1]}_{k}$, $x_i$, $i$, $j$, $u_{i-1}^j(z^{[i-1]}_{k})$, $v_{i-1}^{j, (m)}$, $\mathcal{R}$)
            \EndFor
            \State \parbox[t]{313pt}{ Compute density
            \begin{multline*}
            p_{i}^{(m)}{\left(z^{[i-1]}_{k}\right)} \gets \left\{1-\alpha_{i} + \alpha_{i} \prod_{j=1}^d {c\left(u_{i-1}^{j}(x_{\nr+ k}), v_{i-1}^{j, (m)}; {\rho^j_{i}(z^{[i-1]}_{k})}\right)}\right\} p_{i-1}^{(m)}\left(z^{[i-1]}_{k}\right)
            \end{multline*} \strut}
        \EndFor
        \State $p_{i}(z^{[i-1]}_{k}) \gets \frac{1}{M}\sum_{m=1}^{M} p^{(m)}_{i}(z^{[i-1]}_{k})$ 
        \new{ \State $w^{[i]}_{k} \gets w^{[i-1]}_{k} \cdot p_{i}{\left(z^{[i-1]}_{k}\right)}/p_{i-1}{\left(z^{[i-1]}_{k}\right)}$}
    \EndFor
    \new{ \State\textsc{ess} $\gets \left(\sum_k w^{[i]}_{k}\right)^2/\left(\sum_k w^{[i]^2}_{k}\right)$
    \If {\textsc{ess} $< 0.5\cdot \np$}
        \State $\{z^{[i]}_{k}\}_k \gets $\textsc{resample_with_replacement}$\left(\{z^{[i-1]}_{k}\}_k, \{w^{[i]}_{k}\}_k\right)$
        \State $w^{[i]}_{k} \gets 1$ \textbf{for} $k=1,\ldots,\np$
    \Else 
        \State $z^{[i]}_{k} \gets z^{[i-1]}_{k}$ \textbf{for} $k=1,\ldots,\np$
    \EndIf}
\EndFor\\
\hspace*{4mm} \Return {$\{z^{[i]}_{k}\}_k$} 
\EndProcedure
\end{algorithmic}\label{alg:sample}
\end{algorithm}

\section{Experiments} \label{app:exp}

\subsection{Experimental Details} \label{app:expdetails} %
The UCI data sets \citep{asuncion2007uci} we used are: wine, breast, parkinson (PARKIN), ionosphere (IONO), boston housing (BOSTON), concrete (CONCR), diabetes (DIAB), and digits.

\paragraph{Code} We downloaded the code for MAF and NSF from \url{https://github.com/bayesiains/nsf}, and the code for R-BP from \url{https://github.com/edfong/MP/tree/main/pr_copula}, and implemented EarlyStopping with patience 50, and 200 minimal, and 2000 maximal iterations. Note that we chose the autoregressive version of RQ-NSF over the coupling variant as the former seemed to generally outperform the latter in \cite{durkan2019neural}. The neural network in ARnet-BP was implemented with \texttt{Haiku} \citep{haiku2020github}. The remaining methods are implemented in \texttt{sklearn}. For the DPMM with VI  (mean-field approximation), we use both the diagonal and full covariance function, with default hyperparameters for the priors. The code used to generate these results is available as an additional supplementary directory.

\paragraph{Initialisation} We initialise the predictive densities with a standard normal, the bandwidth parameter with $\rho_0=0.9$, the length scales with $l_2=1, ..., l_{d-1}=1$, and the neural network weights inside ARnet-BP by sampling from a truncated normal with variance proportional to the number of input nodes of the layer. %

\paragraph{Data pre-processing} For each dataset, we standardized each of the attributes by mean-centering and rescaling to have a sample standard deviation of one. Following \cite{papamakarios2017masked}, we eliminated discrete-valued attributes. To avoid issues arising from collinearity, we also eliminated one attribute from each pair of attributes with a Pearson correlation coefficient greater than 0.98.

\paragraph{Hyperparameter tuning} We average over $M=10$ permutations over samples and features. The bandwidth of the \glspl{kde} was found by five-fold cross validation over a grid of 80 log-scale-equidistant values from $\rho=0.1$ to 100. For the \gls{dpmm}, we considered versions with a diagonal (Diag) and full (Full) covariance matrix for each mixture component. We optimized over the weight concentration prior of the \gls{dpmm} by five-fold cross validation with values ranging from $10^{-40}$ to $1$. The model was trained with variational inference using \texttt{sklearn}. The hyperparameters of \glspl{maf} and \glspl{rqnsf} were found with a Bayesian optimisation search. 
For MAF and RQ-NSF, we applied a Bayesian optimisation search over the learning rate $\{3\cdot 10^{-4}, 4\cdot 10^{-4}, 5\cdot 10^{-4}\}$, the batch size $\{512, 1024\}$, the flow steps $\{10, 20\}$, the hidden features $\{256, 512\}$, the number of bins $\{4, 8\}$, the number of transform blocks $\{1, 2\}$ and the dropout probability $\{0, 0.1, 0.2\}$. On each data set, the hyperparameter search ran for more than 5 days. Please see Table \ref{tab:sweep} for the optimal parameters found. For the benchmark UCI data sets, we did not tune the hyperparameters for neither MAF nor RQ-NSF but instead used the standard parameters given by \cite{durkan2019neural}. The kernel parameters of the \gls{gp} are optimised during training, the $\alpha$ resp. $\lambda$ intialization parameter of the linear model over the range from 1 to 2 resp. 0.01 to 0.1, and the hidden layer sizes of the MLP over the values $\{64, 128, 256\}$. 

\begin{table}[ht]
\footnotesize
\centering
\caption{Hyperparameters for MAF and RQ-NSF}

\begin{tabularx}{\textwidth}{p{0.02\textwidth}p{0.18\textwidth}YYYYYYY}%
\toprule
& data &  batch size &  learning rate & flow steps &  hidden nodes &  bins &  transform blocks &  dropout \\
\midrule
\multirow{7}{*}{\rotatebox[origin=c]{90}{MAF}} &  WINE &                   10000 &               0.0003 &                    20 &                    512 &               - &                           1 &                        0.2 \\
&BREAST &                   10000 &               0.0004 &                    20 &                    512 &               - &                           1 &                        0.2 \\
&PARKINSONS &                   10000 &               0.0004 &                    20 &                    512 &               - &                           1 &                        0.2 \\
&IONOSPHERE &                   10000 &               0.0003 &                    20 &                    512 &               - &                           1 &                        0.2 \\
&BOSTON &                   10000 &               0.0003 &                    10 &                    512 &               - &                           1 &                        0.2 \\
&CONCRETE &                    1024 &               0.0003 &                    10 &                    512 &               - &                           1 &                        0.2 \\
&DIABETES &                     10000 &               0.0004 &                    20 &                    512 &               - &                           1 &                        0.2 \\
&CHECKERBOARD &                    10000 &               0.0003 &                    20 &                    512 &               - &                           1 &                    0.2 \\
\midrule
\multirow{8}{*}{\rotatebox[origin=c]{90}{RQ-NSF}} &  WINE &                   10000 &               0.0004 &                    20 &                    512 &               8 &                           1 &                        0.2 \\
&BREAST &                   10000 &               0.0005 &                    10 &                    512 &               8 &                           1 &                        0.2 \\
&PARKINSONS &                   10000 &               0.0005 &                    20 &                    512 &               8 &                           1 &                        0.2 \\
&IONOSPHERE &                   10000 &               0.0003 &                    10 &                    512 &               8 &                           1 &                        0.2 \\
&BOSTON &                   10000 &               0.0003 &                    10 &                    512 &               8 &                           1 &                        0.2 \\
&CONCRETE &                    1024 &               0.0004 &                    20 &                    256 &               8 &                           2 &                        0.1 \\
&DIABETES &                     256 &               0.0004 &                    10 &                    512 &               8 &                           2 &                        0.2 \\
&CHECKERBOARD &                    1024 &               0.0004 &                    10 &                    512 &               8 &                           2 &                        0.1 \\
\bottomrule
\end{tabularx}    
    \label{tab:sweep}
\end{table}

\paragraph{Compute} 
We run all BP and neural network experiments on a single Tesla V100 GPU, as provided in the internal cluster of our department. In total, these experiments required compute of approximately 4000 GPU hours. The remaining experiments were run on a single core of an Intel(R) Xeon(R) Gold 6240 CPU @ 2.60GHz, using up a total of 100 hours.

\subsection{Additional Experimental Results} \label{app:exp-add}
\begin{figure}
    \centering
    \subfloat[AR-BP]{\includegraphics[width = 0.6\textwidth]{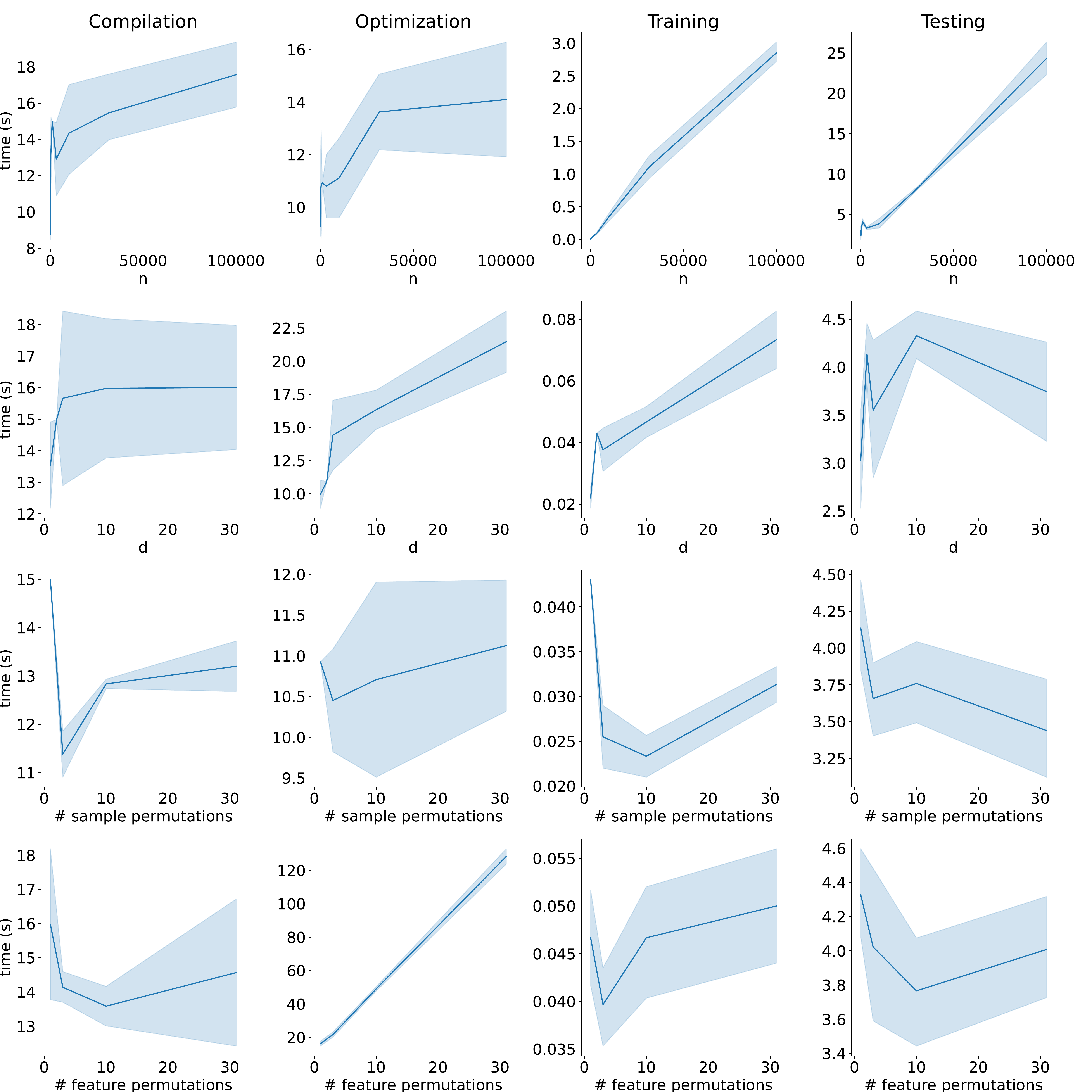}}\\
    \subfloat[ARnet-BP]{\includegraphics[width = 0.6\textwidth]{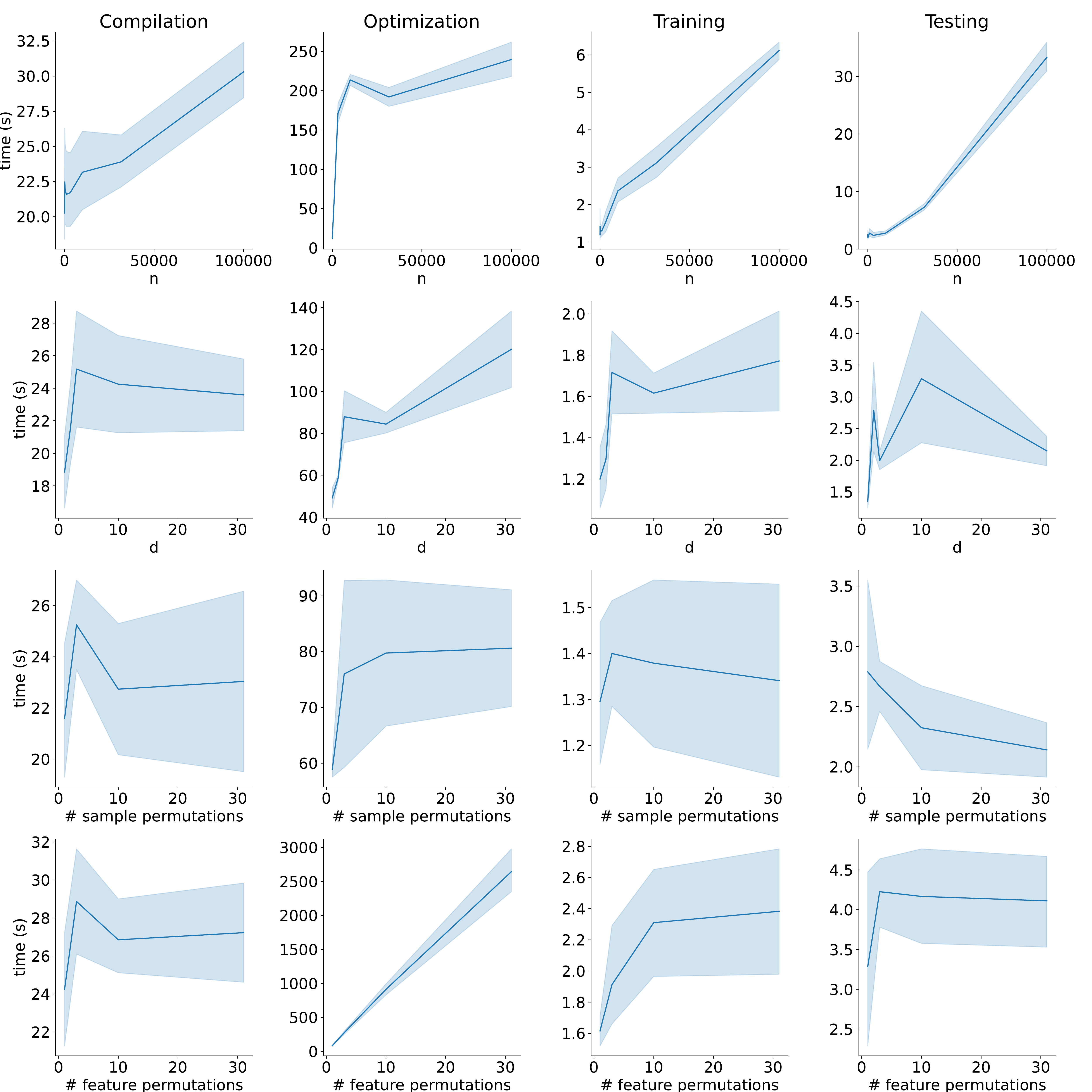}}
    \caption{Computational study: computational time measured in elapsed seconds for a simple \gls{gmm} example. Note that R-BP has the same computational complexity and only saves an indiscernible constant time factor.}
    \label{fig:gmm-comp}
\end{figure}

\paragraph{Computational analysis} \label{app:exp-comp} For the computational study, we consider data sampled from a \gls{gmm}. By default, we set the number of training samples to $n=500$, the number of test samples to $n'=500$, the number of features to $d=2$, the number of mixture components to $K=2$, and the number of feature and samples permutations to 1. In Figure \ref{fig:gmm-comp}, we plot the compute in elapsed seconds w.r.t changes in these parameters. 

\begin{figure}
    \centering
    \includegraphics[width=\textwidth]{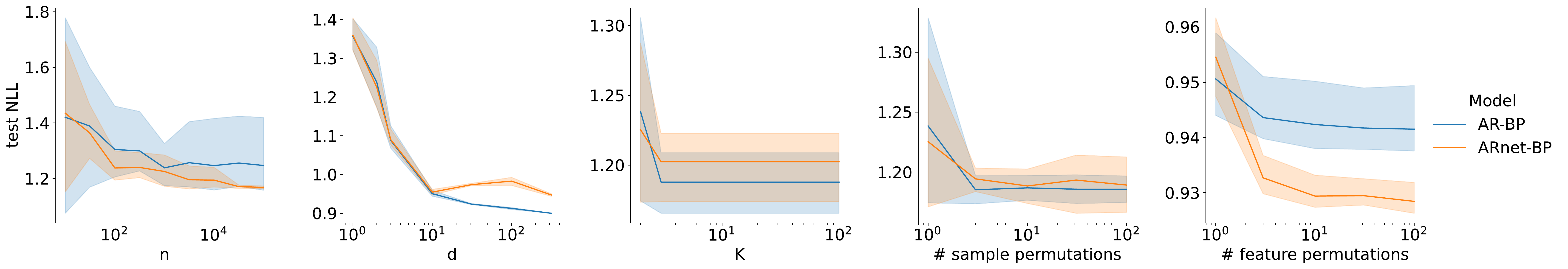}
    \caption{Sensitivity analyis: Average test \gls{nll} over 5 runs reported with standard error for a simple \gls{gmm} example over a range of simulation and parameter settings.}
    \label{fig:gmm-sens}
\end{figure}

\paragraph{Sensitivity analysis} \label{app:exp-sens}
For the sensitivity study, we consider the same simulated \gls{gmm} data as in the computational study, and plot the results in Figure \ref{fig:gmm-sens}. As expected, we observe that the test \gls{nll} decreases in $n$, and in the number of permutations. It also decreases in the number of mixture components. One possible explanation for this is that, as noted by \citet{hahn2018recursive}, R-BP can be interpreted as a mixture of $n$ normal distributions. The \gls{nll} decreases in $d$, as the mixture components are easier to distinguish in higher dimensional covariate spaces.

\paragraph{Ablation study} \label{app:exp-abl}
Figure \ref{fig:gmm-sens} shows the test \gls{nll} of ARnet-BP and AR-BP for the above \gls{gmm} example, as a function of the number of sample permutations, and number of feature permutations. We see that averaging over multiple permutations is crucial to the performance of AR-BP. In Table \ref{tab:smalluci_ablation}, we also show results on the small UCI datasets for: 
\begin{itemize}
    \item a different choice of covariance function, namely a rational quadratic covariance function, defined by $k(x, x_i)=\left(1+\frac{||x-x_i||^2_2}{2\gamma \ell^2}\right)^{-\gamma}$, where $\ell, \gamma>0$ and
    \item a different choice of initial distribution, namely a uniform distribution (unif).
\end{itemize}
\begin{table}[ht]
\caption{Average \gls{nll} with standard error over five runs on five UCI data sets of small-to-moderate size}
\footnotesize
\begin{tabularx}{\textwidth}{p{0.18\textwidth}YYYYY}%
\toprule
{} &                    WINE &                   BREAST &               PARKIN &               IONO &                          BOSTON \\
{n/d} &                89/12 &                   97/14 &               97/16 &               175/30 & 506/13  \\\hline
AR$_d$-BP   &  $\bm{13.22_{\pm 0.04}}$ &   $\bm{6.11_{\pm 0.04}}$&  $\bm{7.21_{\pm 0.12}}$ &    $\bm{16.48_{\pm 0.26}}$ & ${-14.75_{\pm 0.89}}$\\
AR-BP (RQ)       &  ${13.53_{\pm 0.02}}$ &  ${7.39_{\pm 0.06}}$ &  ${8.79_{\pm 0.08}}$ &  ${21.26_{\pm 0.08}}$ &  ${4.49_{\pm 0.00}}$ \\
AR$_{d}$-BP (RQ) &      ${13.36_{\pm 0.04}}$ &     ${6.18_{\pm 0.03}}$ &      ${7.85_{\pm 0.08}}$ &      ${20.25_{\pm 0.09}}$ &   $\bm{-20.41_{\pm 1.28}}$ \\
AR$_d$-BP (unif) &  ${-5.18_{\pm 0.04}}$ &  ${-15.51_{\pm 0.11}}$ &  ${-16.58_{\pm 0.06}}$ &  ${-47.77_{\pm 3.77}}$ &  ${-10.73_{\pm 1.63}}$ \\
\bottomrule
\end{tabularx}
\label{tab:smalluci_ablation}
\end{table}

We observe that none of these ablations consistently outperforms AR$_d$-BP.

\paragraph{Benchmark UCI data sets} \label{app:otherexp}
As we only presented a subset of the results on the benchmark data sets introduced by \cite{papamakarios2017masked} in Section \ref{sec:exp}, we present more results for density estimation on the complete data set in Table \ref{tab:largeuci}. These results underscore that 1) MAF and RQ-NSF outperform any other baseline, the more data is available; 2) KDE underperforms in high-dimensional settings; 3) DPMM is not suitable for every data distribution. Note that evaluation of the R-BP variants take at least 4 days to run on any of the data sets with more than 800,000 observations which is why we omitted those results here. %

\begin{table}[ht]
\caption{Average \gls{nll} with standard error over five runs on benchmark UCI data from \cite{papamakarios2017masked}}
\scriptsize
\begin{tabularx}{\textwidth}{p{0.18\textwidth}YYYYY}%
\toprule
{} &                   POWER &                      GAS &                 HEPMASS &                MINIBOONE &                 BSDS300 \\
n/d & 1,659,917/6 &  852,174/8 & 315,123/21& 29,556/43 & 1,000,000/ 63\\
\midrule
Gaussian    &     ${7.73_{\pm 0.00}}$ &      ${3.59_{\pm 0.00}}$ &        ${27.93_{\pm 0.00}}$ &     ${37.20_{\pm 0.00}}$ &     ${56.45_{\pm 0.00}}$ \\
KDE         &    ${29.39_{\pm 0.00}}$ &  $\bm{-9.61_{\pm 0.00}}$ &     ${26.44_{\pm 0.00}}$ & ${43.88_{\pm 7.52}}$ &    ${63.70_{\pm 10.00}}$ \\
DPMM (Diag) &     ${0.51_{\pm 0.01}}$ &      ${1.20_{\pm 0.02}}$ &        ${25.80_{\pm 0.00}}$ &     ${39.16_{\pm 0.01}}$ &     ${37.55_{\pm 0.02}}$ \\
DPMM (Full) &     ${0.33_{\pm 0.00}}$ &     ${-5.57_{\pm 0.04}}$ &        ${23.40_{\pm 0.02}}$ &     ${18.82_{\pm 0.01}}$ &      ${4.47_{\pm 0.00}}$ \\
MAF       &     ${0.52_{\pm 0.00}}$ &     ${-2.21_{\pm 0.54}}$ &     ${21.10_{\pm 0.04}}$ &     ${12.81_{\pm 0.08}}$ &  ${2.76_{\pm 0.17}}$ \\
RQ-NSF    &  $\bm{0.00_{\pm 0.01}}$ &  ${-6.41_{\pm 0.14}}$ &  $\bm{19.46_{\pm 0.08}}$ &  $\bm{12.51_{\pm 0.19}}$ &  $\bm{2.44_{\pm 0.56}}$ \\
\bottomrule
\end{tabularx}
\label{tab:largeuci}
\end{table}

\paragraph{Image examples} \label{app:exp-image}
We provide preliminary results on two image datasets, digits and MNIST, in Table \ref{tab:images}. Note that the AR-BP copula updates investigated here were not designed with computer vision tasks in mind. The rich parameterization allows the model to overfit to the data leading to a prequential negative log-likelihood of at least -684 at train time while the test \gls{nll} is considerably higher. ARnet-BP, on the other hand, helps to model the complex data structure more efficiently. We expect that further extensions based on, for instance, convolutional covariance functions \citep{van2017convolutional} may prove fruitful. 
\vspace{-2mm}
\begin{figure}
\centering
\footnotesize
\subfloat[True data]{\includegraphics[width = \textwidth/5]{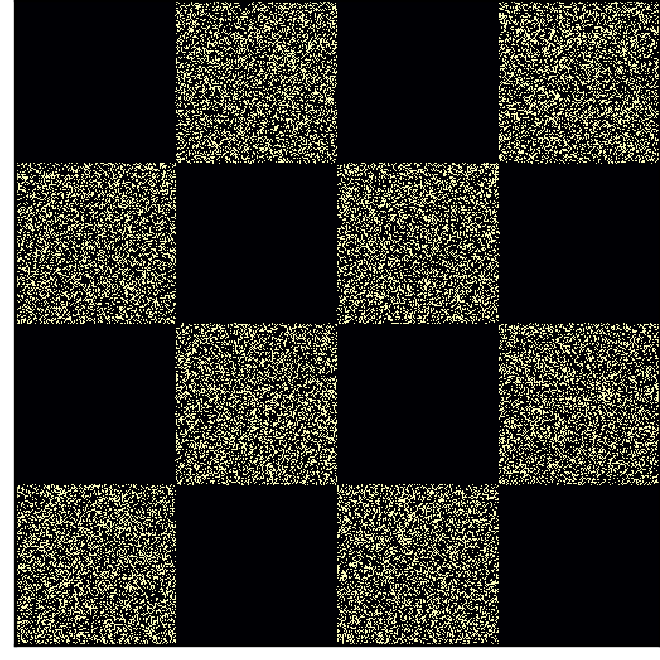}} 
\subfloat[R-BP]{\includegraphics[width = \textwidth/5]{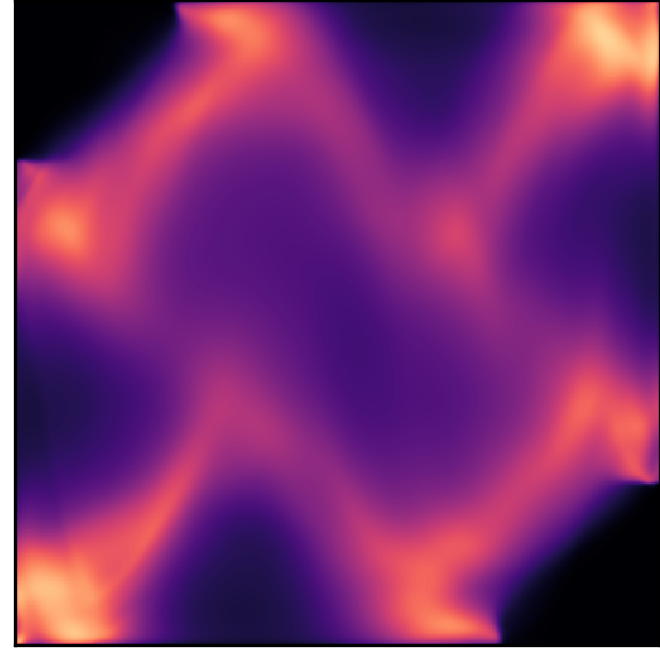}}
\subfloat[R$_{d}$-BP]{\includegraphics[width = \textwidth/5]{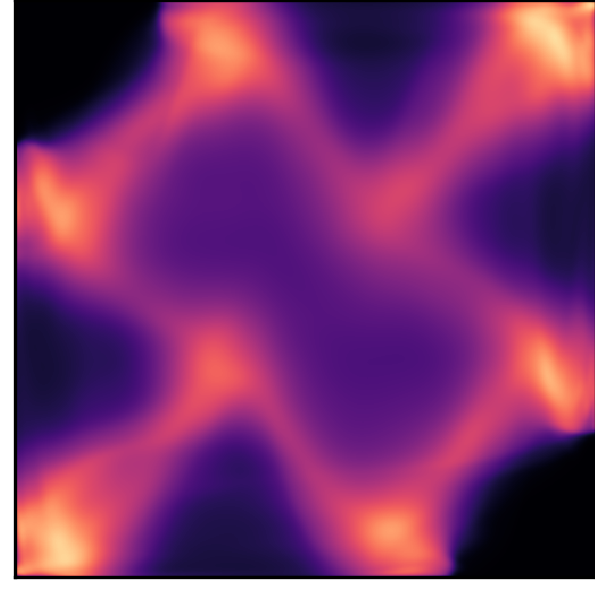}}
\subfloat[AR-BP]{\includegraphics[width = \textwidth/5]{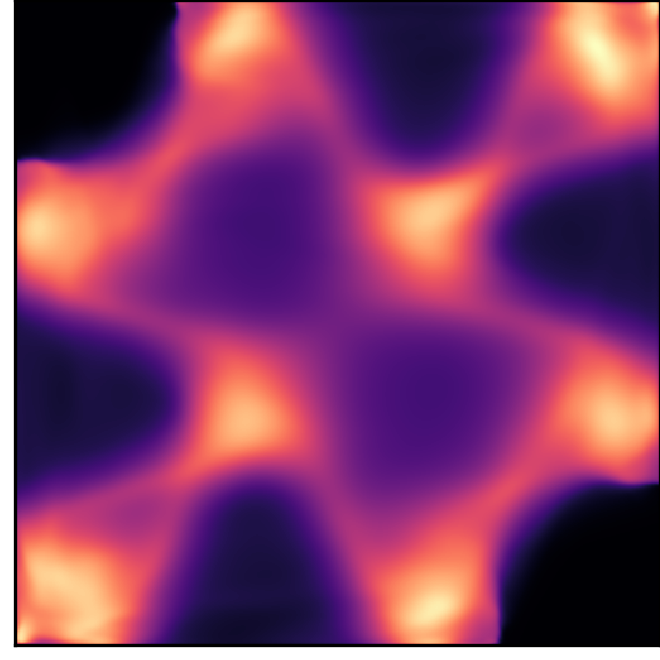}}\\
\subfloat[AR$_{d}$-BP]{\includegraphics[width = \textwidth/5]{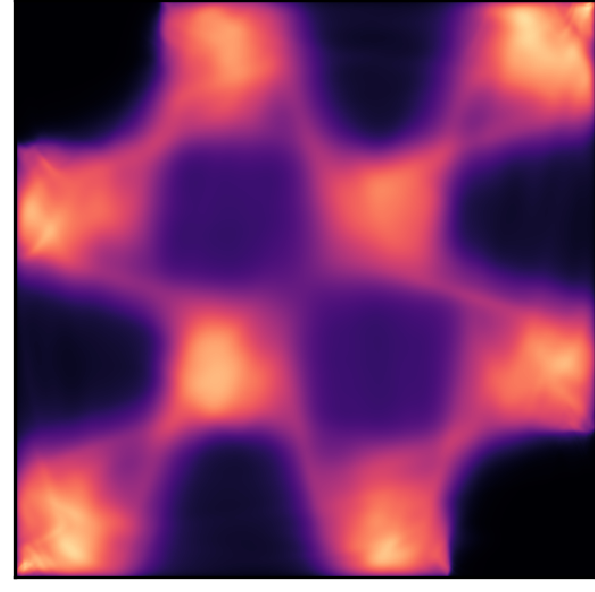}}
\subfloat[ARnet-BP]{\includegraphics[width = \textwidth/5]{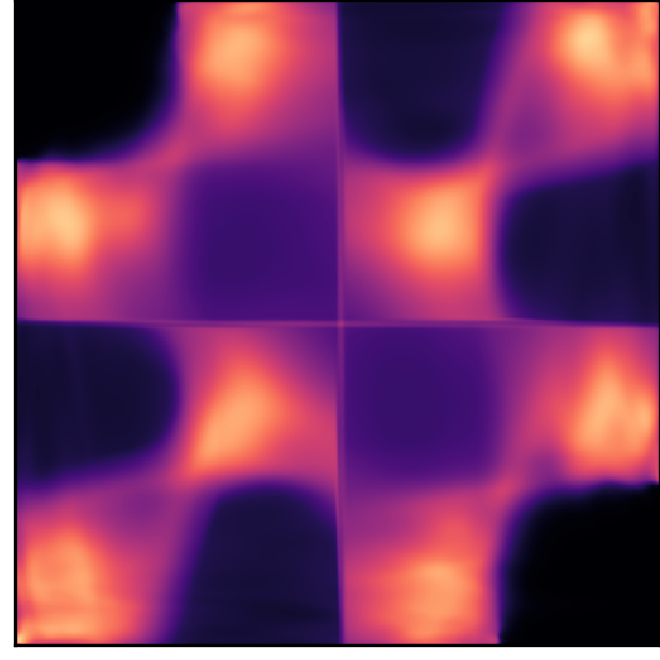}} 
\subfloat[MAF]{\includegraphics[width = \textwidth/5]{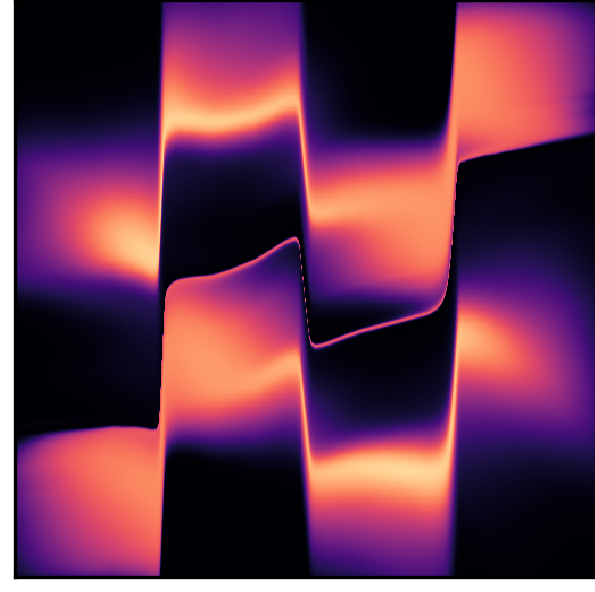}}
\subfloat[RQ-NSF]{\includegraphics[width = \textwidth/5]{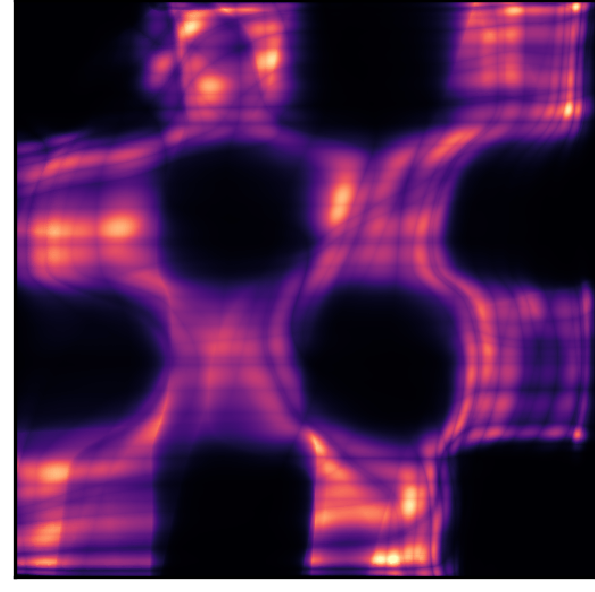}}
\caption{Scatter plot and density estimates of 60,000 observations sampled from a chessboard data distribution. Test log likelihoods are R-BP: $2.25_{\pm 0.0}$, R$_{d}$-BP : $2.19_{\pm 0.0}$, AR-BP: $2.21_{\pm 0.0}$, AR$_{d}$-BP: $2.10_{\pm 0.0}$, ARnet BP : $2.19_{\pm 0.0}$, MAF : $2.09_{\pm 0.0}$, RQ-NSF : $2.05_{\pm 0.0}$.}
\label{fig:chess-full}
\end{figure}
\begin{table}[ht]
\caption{Image datasets: average test \gls{nll} over five runs displayed with standard error}
\scriptsize
\centering\begin{tabular}{lll}
\toprule
{} &                   DIGITS &                    MNIST \\
\midrule
MAF       &     $\bm{-8.76_{\pm 0.10}}$ &     ${-7.14_{\pm 0.48}}$ \\
RQ-NSF    &     ${-6.17_{\pm 0.13}}$ &  ${-8.49_{\pm 0.03}}$ \\
R-BP      &     ${-8.80_{\pm 0.00}}$ &     ${-9.04_{\pm 0.07}}$ \\
R$_d$-BP  &     ${-7.46_{\pm 0.12}}$ &     ${-7.73_{\pm 0.07}}$ \\
AR-BP     &     ${-8.66_{\pm 0.03}}$ &    $-{7.31_{\pm 42.54}}$ \\
AR$_d$-BP &     ${-7.46_{\pm 0.18}}$ &    $-{8.32_{\pm 61.92}}$ \\
ARnet-BP  &     ${-7.72_{\pm 0.28}}$ &    $\bm{-9.20_{\pm 0.10}}$ \\
\bottomrule
\end{tabular}
\label{tab:images}
\vspace{-6mm}
\end{table}

\begin{figure}
    \centering
    \includegraphics[width=0.7\textwidth]{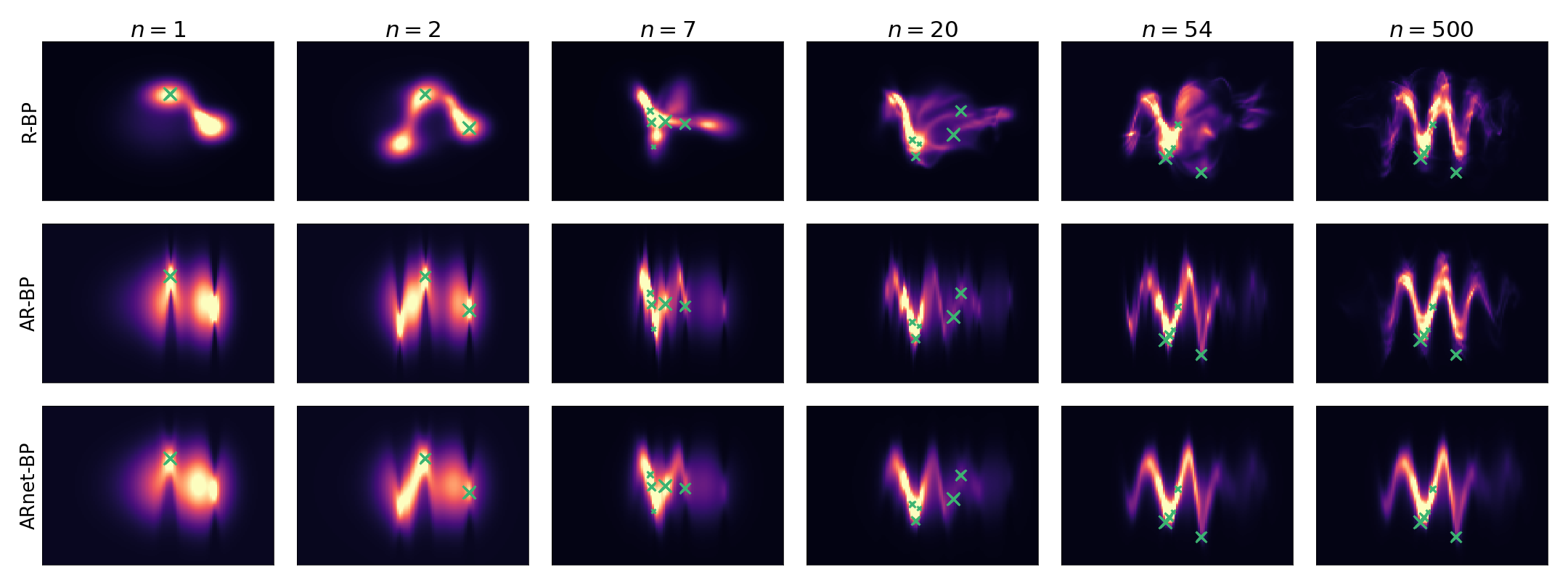}
    \caption{Illustration of the importance of an autoregressive kernel. We trained the models on 500 data points sampled according to a sine wave distribution (given in Figure \ref{fig:toys}). We visualise the predictive density after observing a different number, $\nr$, of observations, highlighting the last five points with \crossmid. We observe that for highly non-linear relationships between $x^1$ and $x^2$, the optimal bandwidth of R-BP is quite high ($\rho=0.93$) which results in strong overfitting. Even when we choose $\rho_0=0.93$ for AR-BP and ARnet-BP, we observe that these models learn the true data distribution with fewer samples than R-BP does.}
    \label{fig:sinewave}
\end{figure}

\begin{figure}
\centering
\subfloat{\includegraphics[width = 0.6in]{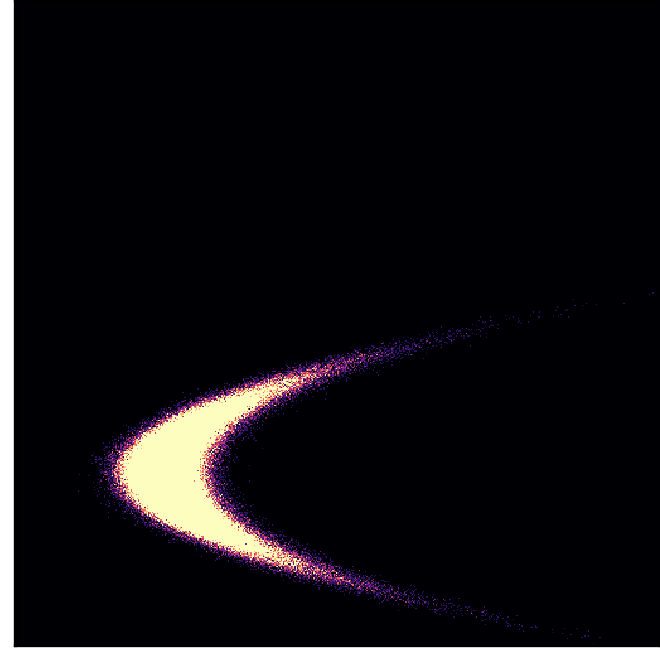}} 
\subfloat{\includegraphics[width = 0.6in]{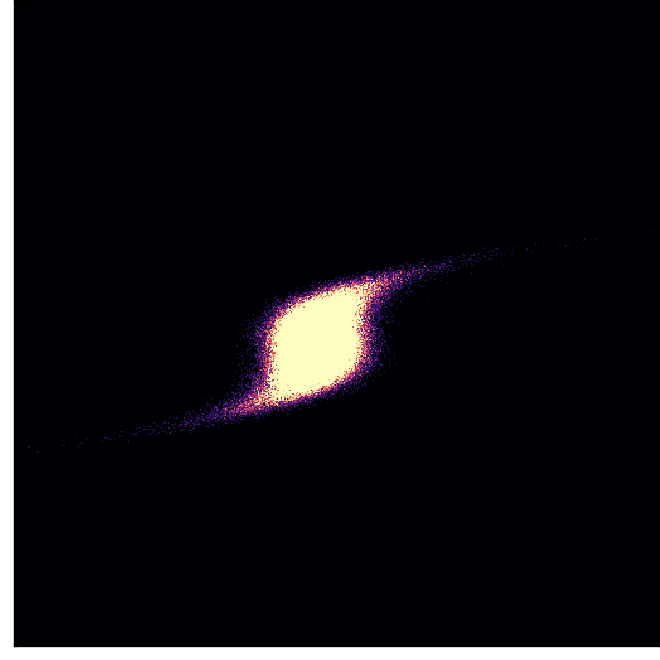}}
\subfloat{\includegraphics[width = 0.6in]{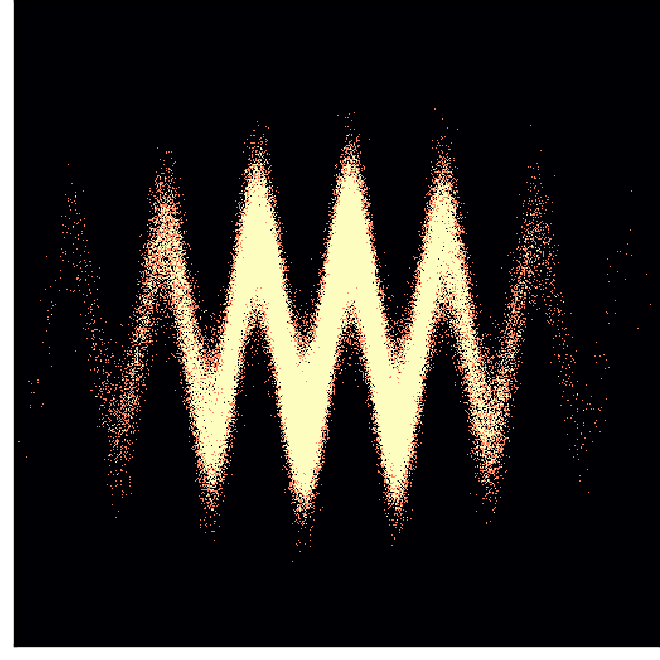}}
\subfloat{\includegraphics[width = 0.6in]{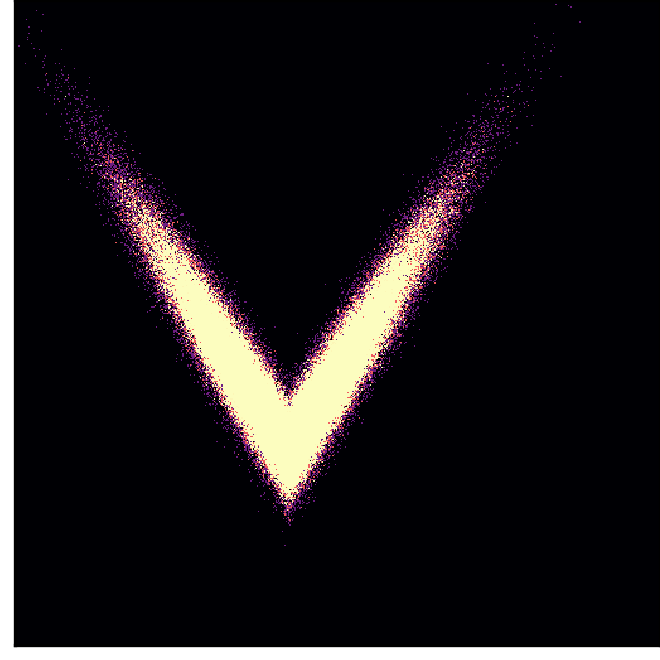}} 
\subfloat{\includegraphics[width = 0.6in]{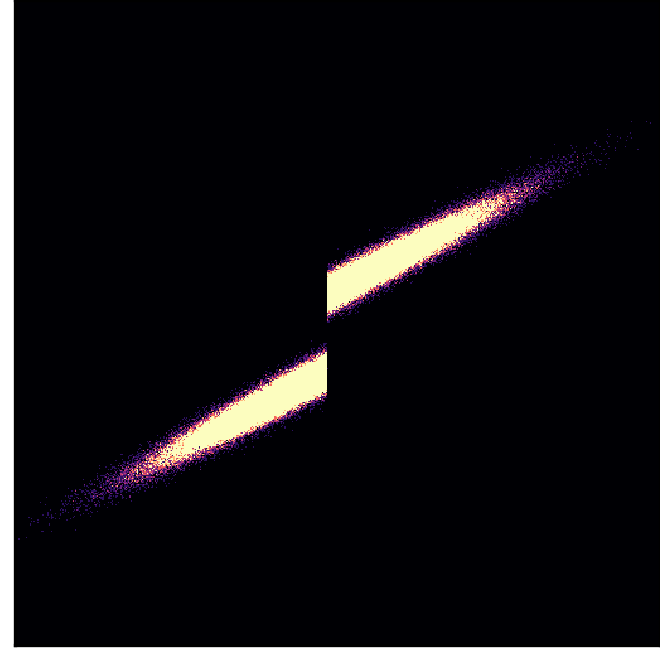}} \\

\subfloat{\includegraphics[width = 0.6in]{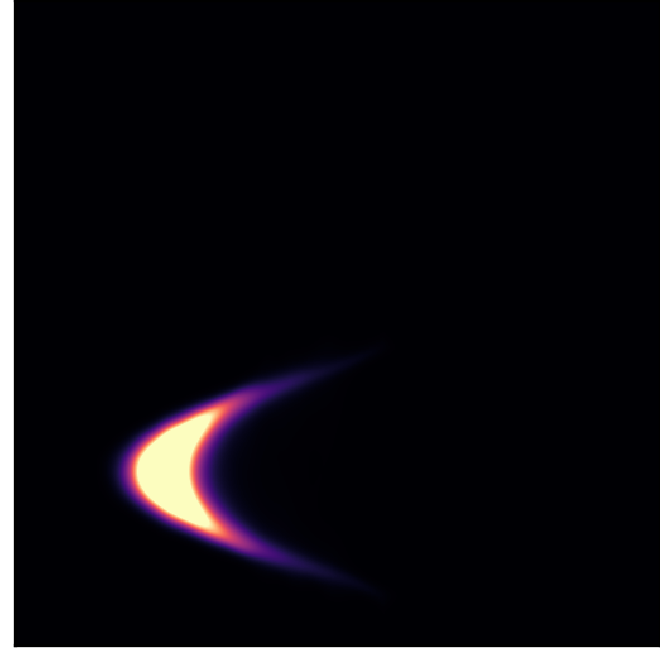}}
\subfloat{\includegraphics[width = 0.6in]{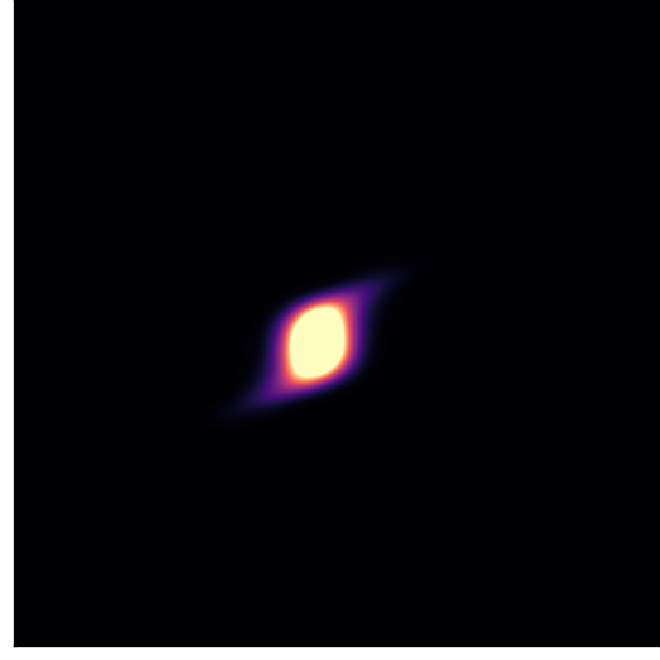}}
\subfloat{\includegraphics[width = 0.6in]{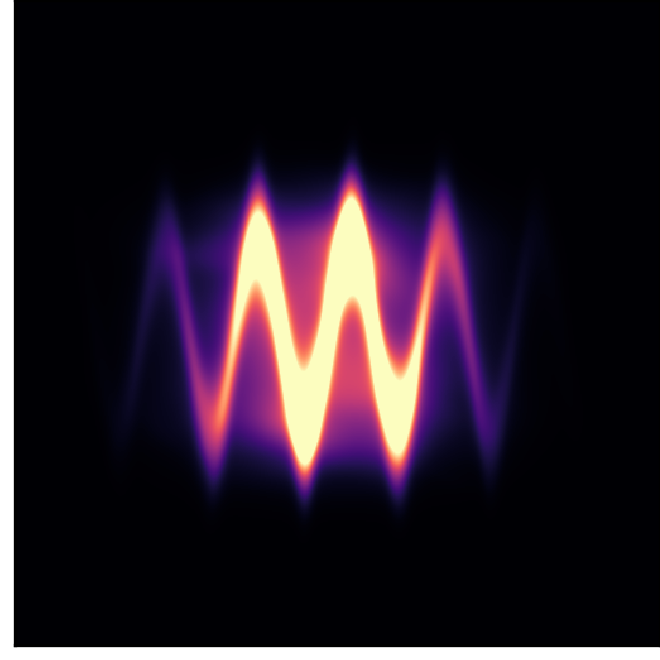}}
\subfloat{\includegraphics[width = 0.6in]{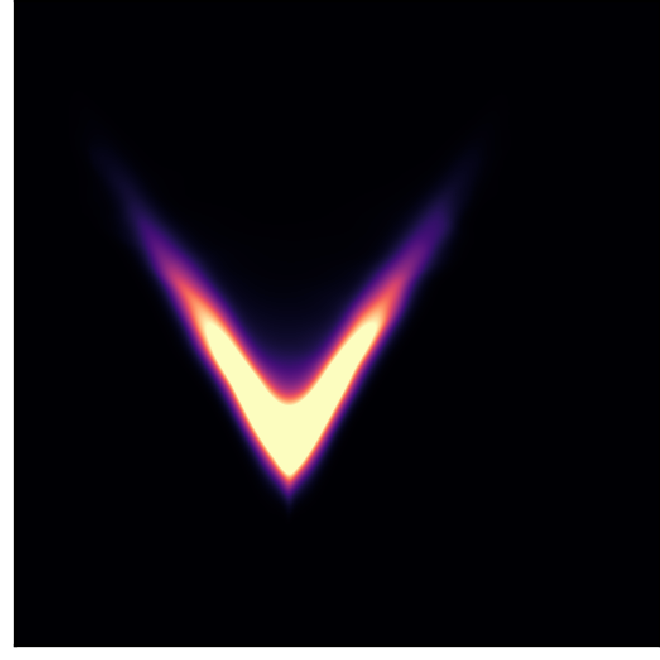}} 
\subfloat{\includegraphics[width = 0.6in]{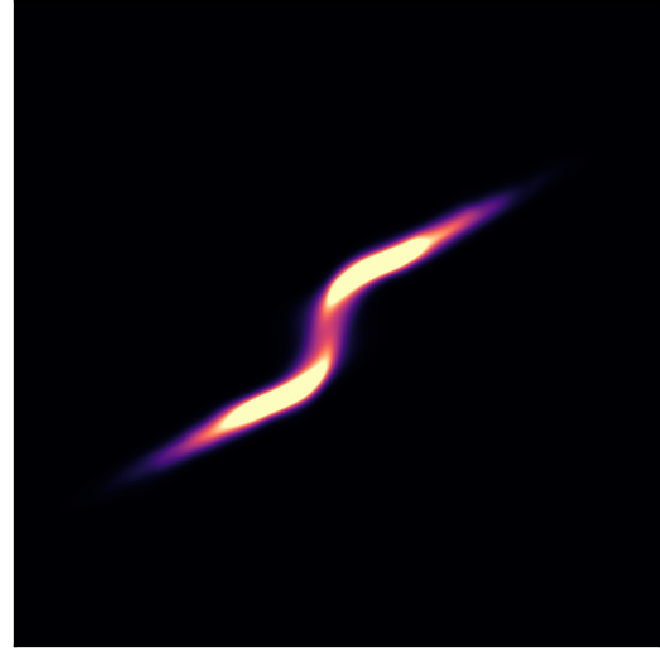}} 

\caption{Scatter plots of 60,000 samples from different data distributions in the first row, and corresponding autoregressive predictive density estimates in the second row.}
\label{fig:toys}
\end{figure}

\paragraph{Toy examples} \label{app:exp-toy}
Figure \ref{fig:chess-full} shows density estimates for the introductory example of the checkerboard distribution in a large data regime. We observe that neural-network-based methods outperform the AR-BP alternatives. Nevertheless, AR-BP performs better than the baseline R-BP. An illustration of this behaviour on another toy example is also given in Figure \ref{fig:sinewave}. Figure \ref{fig:toys} shows density estimates from AR-BP on a number of complex distributions.


\end{document}